\documentclass[10pt,twocolumn,letterpaper]{article}

\usepackage[pagenumbers]{cvpr}              

\usepackage{graphicx}
\usepackage{amsmath}
\usepackage{mathtools}
\usepackage{amssymb}
\usepackage{booktabs}
\usepackage{dashrule}
\usepackage{layouts}
\usepackage{multirow}
\usepackage{tabularx}
\usepackage{pbox}
\usepackage{comment}
\usepackage{setspace}
\makeatletter
\@namedef{ver@everyshi.sty}{}
\makeatother
\usepackage{pgf}
\usepackage[hang,flushmargin]{footmisc}
\usepackage{anyfontsize}
\graphicspath{{figs/}}

\usepackage[pagebackref,breaklinks,colorlinks]{hyperref}

\usepackage[capitalize]{cleveref}
\crefname{section}{Sec.}{Secs.}
\Crefname{section}{Section}{Sections}
\Crefname{table}{Table}{Tables}
\crefname{table}{Tab.}{Tabs.}

\def\cvprPaperID{####} 
\def\confName{CVPR}
\def\confYear{2023}

\usepackage{overpic}
\usepackage{enumitem} 
\usepackage{overpic} 
\usepackage{color}
\usepackage{soul} 

\definecolor{dark-gray}{gray}{0.25}
\definecolor{myorange}{RGB}{247, 148, 2}
\definecolor{myorange2}{RGB}{200,92,18}
\definecolor{myblue}{RGB}{42,93,166}
\definecolor{mypink}{RGB}{203, 56, 255}

\begin{document}

\title{DINER: Depth-aware Image-based NEural Radiance fields\vspace{-0.2cm}}

\author{
Malte Prinzler$^{1,3}$\\
{\tt\small malte.prinzler@tuebingen.mpg.de}
\and
Otmar Hilliges$^{2}$\\
{\tt\small otmar.hilliges@inf.ethz.ch}
\vspace{0.1cm}
\and
Justus Thies$^{1}$\\
{\tt\small justus.thies@tuebingen.mpg.de}
\and
\vspace{-0.2cm}
\\
$^1$Max Planck Institute for Intelligent Systems, Tübingen, Germany~~~~\\
$^2$ETH Zürich~~~~
$^3$Max Planck ETH Center for Learning Systems
}

\twocolumn[{%
\renewcommand\twocolumn[1][]{#1}%
\vspace{-1cm}
\maketitle
\begin{center}
    \centering
    \captionsetup{type=figure}
    \vspace{-1cm}
    \def\svgwidth{.9\linewidth}
    \hspace*{-.5cm}\mbox{
\begingroup%
  \makeatletter%
  \providecommand\color[2][]{%
    \errmessage{(Inkscape) Color is used for the text in Inkscape, but the package 'color.sty' is not loaded}%
    \renewcommand\color[2][]{}%
  }%
  \providecommand\transparent[1]{%
    \errmessage{(Inkscape) Transparency is used (non-zero) for the text in Inkscape, but the package 'transparent.sty' is not loaded}%
    \renewcommand\transparent[1]{}%
  }%
  \providecommand\rotatebox[2]{#2}%
  \newcommand*\fsize{\dimexpr\f@size pt\relax}%
  \newcommand*\lineheight[1]{\fontsize{\fsize}{#1\fsize}\selectfont}%
  \ifx\svgwidth\undefined%
    \setlength{\unitlength}{960bp}%
    \ifx\svgscale\undefined%
      \relax%
    \else%
      \setlength{\unitlength}{\unitlength * \real{\svgscale}}%
    \fi%
  \else%
    \setlength{\unitlength}{\svgwidth}%
  \fi%
  \global\let\svgwidth\undefined%
  \global\let\svgscale\undefined%
  \makeatother%
  \begin{picture}(1,0.29537499)%
    \lineheight{1}%
    \setlength\tabcolsep{0pt}%
    \put(0,0){\includegraphics[width=\unitlength,page=1]{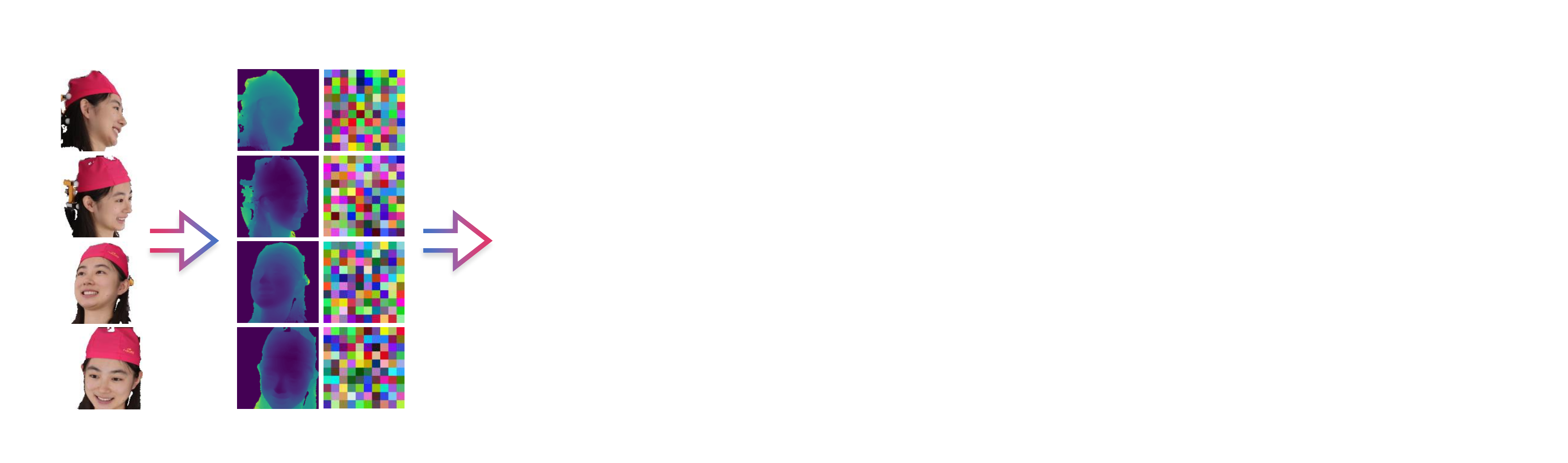}}%
    \put(0,0){\includegraphics[width=\unitlength,page=1]{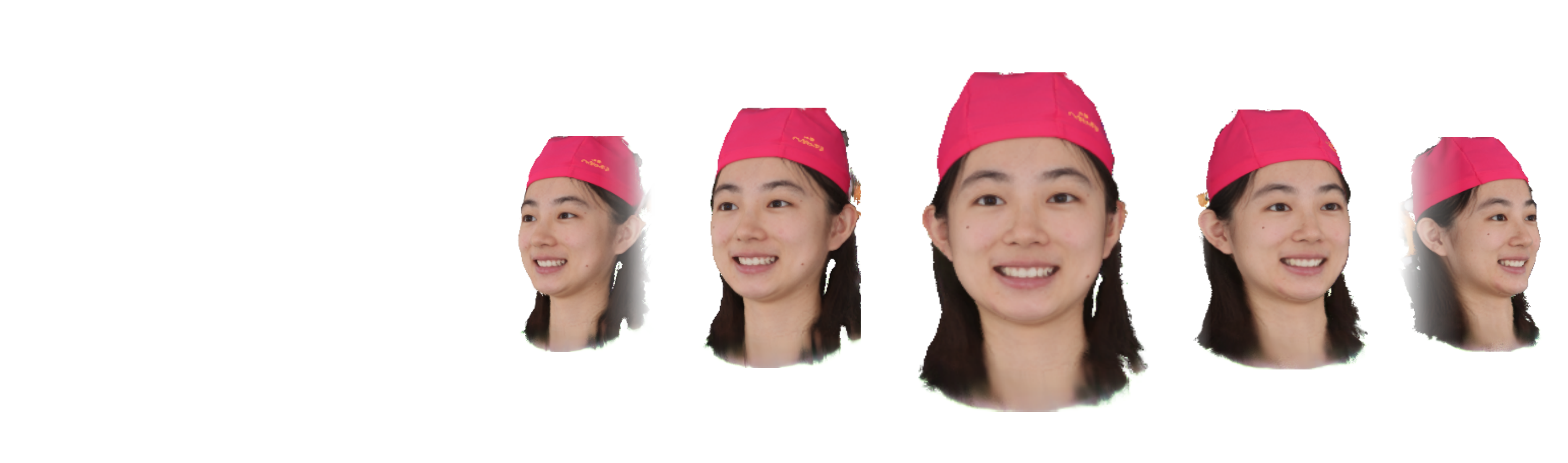}}%
    \footnotesize
    \put(0.06421449,0.01){\color[rgb]{0,0,0}\makebox(0,0)[t]{\lineheight{1.25}\smash{\begin{tabular}[t]{c}4 Input Views\end{tabular}}}}%
    \put(0.205,0.01){\color[rgb]{0,0,0}\makebox(0,0)[t]{\lineheight{1.25}\smash{\begin{tabular}[t]{c}Depth + Features\end{tabular}}}}%
    \put(0.66,0.01){\color[rgb]{0,0,0}\makebox(0,0)[t]{\lineheight{1.25}\smash{\begin{tabular}[t]{c}Volumetric Reconstruction\end{tabular}}}}%
  \end{picture}%
\endgroup%
}\hspace{-.5cm}
      \caption{
      Based on sparse input views, we predict depth and feature maps to infer a volumetric scene representation in terms of a radiance field which enables novel viewpoint synthesis.
      The depth information allows us to use input views with high relative distance such that the scene can be captured more completely and with higher synthesis quality compared to previous state-of-the-art methods.
      %
      }
      \label{fig:teaser}
\end{center}%
}]

\begin{abstract}
We present Depth-aware Image-based NEural Radiance fields (DINER). 
Given a sparse set of RGB input views, we predict depth and feature maps to guide the reconstruction of a volumetric scene representation that allows us to render 3D objects under novel views. 
Specifically, we propose novel techniques to incorporate depth information into feature fusion and efficient scene sampling. 
In comparison to the previous state of the art, DINER achieves higher synthesis quality and can process input views with greater disparity. 
This allows us to capture scenes more completely without changing capturing hardware requirements and ultimately enables larger viewpoint changes during novel view synthesis. 
We evaluate our method by synthesizing novel views, both for human heads and for general objects, and observe significantly improved qualitative results and increased perceptual metrics compared to the previous state of the art. 
The code is publicly available through the \href{https://malteprinzler.github.io/projects/diner/diner.html}{Project Webpage}.
\end{abstract}

%
\section{Introduction}
\label{sec:intro}
%
%


%
In the past few years, we have seen immense progress in digitizing humans for virtual and augmented reality applications.
Especially with the introduction of neural rendering and neural scene representations~\cite{tewari2020neuralrendering, tewari2022advances}, we see 3D digital humans that can be rendered under novel views while being controlled via face and body tracking~\cite{Zheng2021, Gafni2020, authenticvolumetricavatars, lombardi2021mixture, wang2021hvh, rosu2022neuralstrands, DBLP:journals/corr/abs-2112-01554,weng2022humannerf, peng2021neuralbody, Grigorev_2021_CVPR_stylepeople, ARAH:ECCV:2022}.
Another line of research reproduces general 3D objects from few input images without aiming for control over expressions and poses \cite{wang2021ibrnet, chen:mvsnerf, yu2020pixelnerf, Mihajlovic:ECCV2022, suhail2022generalizable, srt22}. 
We argue that this offers significant advantages in real-world applications like video-conferencing with holographic displays: 
(i) it is not limited to heads and bodies but can also reproduce objects that humans interact with, (ii) even for unseen extreme expressions, fine texture details can be synthesized since they can be transferred from the input images, (iii) only little capturing hardware is required e.g. four webcams suffice, and (iv) the approach can generalize across identities such that new participants could join the conference ad hoc without requiring subject-specific optimization.
%
%
Because of these advantages, we study the scenario of reconstructing a volumetric scene representation for novel view synthesis from sparse camera views.
Specifically, we assume an input of four cameras with high relative distances to observe large parts of the scene.
Based on these images, we condition a neural radiance field~\cite{mildenhall2020nerf} which can be rendered under novel views including view-dependent effects. We refer to this approach as \emph{image-based neural radiance fields}.
%
%
%
It implicitly requires estimating the scene geometry from the source images. 
However, we observe that even for current state-of-the-art methods the geometry estimation often fails and significant synthesis artifacts occur when the distance between the source cameras becomes large because they rely on implicit correspondence search between the different views. 
%
%
Recent research demonstrates the benefits of exploiting triangulated landmarks to guide the correspondence search \cite{Mihajlovic:ECCV2022}.
However, landmarks have several drawbacks: They only provide sparse guidance, are limited to specific classes, and the downstream task is bounded by the quality of the keypoint estimation, which is known to deteriorate for profile views.
To this end, we propose \emph{DINER} to compute an image-based neural radiance field that is guided by estimated dense depth.
This has significant advantages: depth maps are not restricted to specific object categories, provide dense guidance, and are easy to attain via either a commodity depth sensor or off-the-shelf depth estimation methods. 
Specifically, we leverage a state-of-the-art depth estimation network~\cite{ding2022transmvsnet} to predict depth maps for each of the source views and employ an encoder network that regresses pixel-aligned feature maps. 
DINER exploits the depth maps in two important ways: (i) we condition the neural radiance field on the deviation between sample location and depth estimates which provides strong prior information about visual opacity, and (ii) we focus sampling on the estimated surfaces to improve sampling efficiency. 
%
Furthermore, we improve the extrapolation capabilities of image-based NeRFs by padding and positionally encoding the input images before applying the feature extractor. 
Our model is trained on many different scenes and at inference time, four input images suffice to reconstruct the target scene in one inference step.  
As a result, compared to the previous state of the art, \emph{DINER} can reconstruct 3D scenes from more distinct source views with better visual quality, while allowing for larger viewpoint changes during novel view synthesis. 
We evaluate our method on the large-scale FaceScape dataset \cite{yang2020facescape} on the task of novel view synthesis for human heads from only four highly diverse source views and on general objects in the DTU dataset~\cite{jensen2014large}.
For both datasets, our model outperforms all baselines by a significant margin.

\medskip
\noindent
In summary, \emph{DINER} is a novel method that produces volumetric scene reconstructions from few source views with higher quality and completeness than the previous state of the art. In summary, we contribute:
\vspace{-0.2cm}
\begin{itemize}
    \setlength\itemsep{0em}
    \item an effective approach to condition image-based NeRFs on depth maps predicted from the RGB input, 
    \item a novel depth-guided sampling strategy that increases efficiency,
    \item and a method to improve the extrapolation capabilities of image-based NeRFs by padding and positionally encoding the source images prior to feature extraction.
\end{itemize}

\begin{figure*}[t!]
    \centering
    \def\svgwidth{\linewidth}
    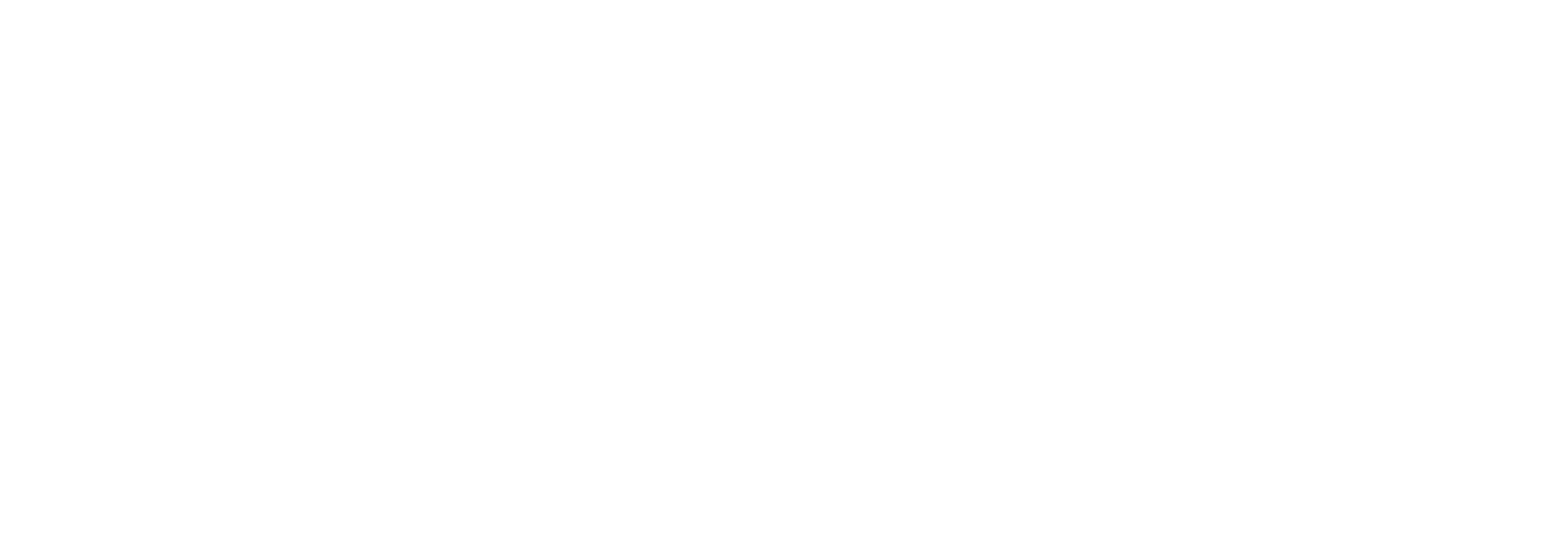
    \caption{Method overview. Given few source images, we first regress depth and feature maps. Image padding and positional encoding prior to the feature map prediction improve extrapolation beyond the source image frustums (\Cref{sec:feature_extrapolation}). Sampled points along target rays are projected onto the source camera planes to interpolate feature vectors $\boldsymbol{\omega}^{(i)}$ and the deviations $\Delta^{(i)}_z$ between the sampling point and predicted depth. They are processed into view-wise intermediate feature vectors $\boldsymbol{V}^{(i)}$ (\Cref{sec:depth_awareness}). The average-pooled intermediate feature vectors of all source views determine the color $\boldsymbol{c}$ and opacity $\sigma$ of each sampling point. The final colors of the target rays are obtained through standard volumetric rendering~\eqref{eq:volumetric_rendering_sum}. Depth-guided sampling increases sampling efficiency (\Cref{sec:depth_guided_sampling}).
}
    \label{fig:method}
\end{figure*}

\section{Related Work}
\label{sec:related}
Our work is related to recent approaches on 3D head avatar generation, and to general neural radiance fields that are reconstructed from a sparse set of input images.
%

\vspace{-0.2cm}
\paragraph{Neural Radiance Fields}
Neural radiance fields (NeRF) \cite{mildenhall2020nerf} and their derivatives have become a popular choice for representing photo-realistic 3D scenes, both for static \cite{Barron2021,tancik2022blocknerf,Mildenhall2022CVPR,mueller2022instant, Chan2021, Xu2022, rematas2021urban, Efficientnet} and dynamic \cite{Pumarola2020,Gafni2020,Park2021, kania2021conerf,park2021nerfies} cases. 
While originally requiring many training images per scene and long training and inference times, latest research focused on increasing data sufficiency \cite{Niemeyer2021Regnerf, roessle2022depthpriorsnerf, rematas2021urban, kangle2021dsnerf} as well as making NeRFs more efficient and faster \cite{Chan2021, chen2022tensorf, mueller2022instant, Hu2022, fastnerf}. Some of these methods even exploit depth information \cite{roessle2022depthpriorsnerf, rematas2021urban, kangle2021dsnerf}, however, they employ it during scene-specific optimization.
%

\vspace{-0.2cm}
\paragraph{Image-Based Rendering}
The above methods require scene-specific optimization which is time-consuming and expensive.
Methods that extract features from as little as one source image and warp them for novel view synthesis \cite{Siarohin2019, Zakharov20, megaportraits, wang2021facevid2vid} offer a fast and efficient alternative but produce artifacts for strong rotations.
Image-based NeRFs \cite{yu2020pixelnerf, wang2021ibrnet, Mihajlovic:ECCV2022, chen:mvsnerf} solve such artifacts through explicit 3D reasoning by using few source images to reconstruct NeRF-like representations in one inference step.
However, for source images with small overlap artifacts occur.  
Triangulated landmarks may be used to guide the geometry estimation \cite{Mihajlovic:ECCV2022}, but they only offer sparse guidance and cannot be applied to arbitrary objects. 
Our method instead exploits depth maps which are more general and provide denser guidance.

\vspace{-0.2cm}
\paragraph{Head Avatars}
The research on head avatar synthesis can be separated along many dimensions~\cite{zollhoefer2018facestar,tewari2022advances}. 
With regard to the choice of representation, the different approaches either rely on image-to-image translation \cite{buehler2021varitex, kim2018deep, thies2019deferred, Zakharov20, Siarohin2019}, volumetric rendering \cite{Gafni2020, lombardi2021mixture, voltemorph, headnerf, wang2021hvh, rosu2022neuralstrands, Wang_2021_CVPR}, or non-transparent surface textures \cite{Zheng2021, DBLP:journals/corr/abs-2112-01554, oneshotmeshavatars, DECA:Siggraph2021, Ramon_2021_ICCV, authenticvolumetricavatars}.
The methods reconstruct head avatars from as little as one source image \cite{Zakharov20, Siarohin2019, oneshotmeshavatars, buehler2021varitex, wang2021facevid2vid, megaportraits, headnerf}, monocular videos \cite{authenticvolumetricavatars, DBLP:journals/corr/abs-2112-01554, Gafni2020, kim2018deep}, or entire camera domes \cite{lombardi2021mixture, wang2021hvh,voltemorph, Lombardi:2019, Wang_2021_CVPR}.
Our method relies on four source images to reconstruct a volumetric head representation, allowing us to synthesize even thin semitransparent structures like hair plausibly. 
The focus of our method is not to learn an animatable avatar, instead, we focus on reconstructing a high-quality volumetric representation from few (e.g., four) and widely spaced source views to enable light-weight capture of immersive, viewpoint-independent video (e.g., for 3D telepresence applications).
Furthermore, our approach is not limited to human heads but can be applied to general objects as well.
\section{Background}
\label{sec:background}
Before detailing the architecture of our pipeline, we briefly review the concepts which it is based on, namely NeRFs in general and image-based NeRFs in particular, and introduce the used notation.


\paragraph{NeRF}
Neural radiance fields (NeRF)~\cite{mildenhall2020nerf} employ multi-layer perceptrons (MLP) to implicitly parameterize a continuous volumetric function $f$ that maps a 3D position $\boldsymbol{x}$ and a view direction vector $\boldsymbol{d}$ to a view-dependent color value $\boldsymbol{c}$ and an isotropic optical density $\sigma$, so that $(\boldsymbol{c}, \sigma) = f(\boldsymbol{x}, \boldsymbol{d})$.
To render the scene, rays are cast into the scene for every pixel of the target image.
We assume such a ray is parametrized by $\boldsymbol{r}(t) = \boldsymbol{o} + t \cdot \boldsymbol{d}$ with near and far plane $t_\text{near}, t_\text{far}$ respectively.
NeRF samples 3D points along the ray $t_j \sim [t_\text{near},t_\text{far}]$, estimates their color and optical density $(\boldsymbol{c}_j, \sigma_j) = f(\boldsymbol{r}(t_j), \boldsymbol{d})$, and integrates the results along the ray following volumetric rendering:
%
\begin{equation}
    \scriptsize
    \boldsymbol{\hat{C}}(\boldsymbol{r}) = \sum_{i=j}^{N} T_j (1-\exp(-\sigma_j\delta_j))\boldsymbol{c_j} 
    \text{~with} ~ T_j = \exp\left(-\sum^{j-1}_{k=1} \sigma_k\delta_k\right) ,
\label{eq:volumetric_rendering_sum}
\end{equation}
where $\delta_j = t_{j+1} - t_j$ denotes the distance between adjacent samples.
In practice, NeRF employs a coarse-to-fine strategy to place samples more efficiently: a coarse MLP is queried to determine regions of importance around which the sample density is increased for evaluating a fine MLP that regresses the final ray color.



\paragraph{Image-Based NeRFs}
Image-based NeRFs enable generalization across scenes by conditioning the NeRF on features extracted from source images. We build our model on top of the pixelNeRF~\cite{yu2020pixelnerf} pipeline.
Assuming $N$ source images $\{\boldsymbol{I}^{(i)}\}_{i=1}^N$ with known extrinsics $\boldsymbol{P}^{(i)}=[\boldsymbol{R}^{(i)}~\boldsymbol{t}^{(i)}]$ and intrinsics $\boldsymbol{K}^{(i)}$, a 2D-convolutional encoding network $\mathcal{E}$ extracts feature maps $\boldsymbol{W}^{(i)}$ for each source image:
\begin{equation}
    \boldsymbol{W}^{(i)}=\mathcal{E}(\boldsymbol{I}^{(i)}). 
\end{equation}
For obtaining the color and opacity of a point, its 3D position $\boldsymbol{x}$ and view direction $\boldsymbol{d}$ are first transformed into the source view coordinate systems:
\begin{equation}
        \boldsymbol{x}^{(i)} = \boldsymbol{P}^{(i)}  \circ \boldsymbol{x}, \quad \boldsymbol{d}^{(i)} = \boldsymbol{R}^{(i)}  \circ \boldsymbol{d} ,
\end{equation}
after which $\boldsymbol{x}^{(i)}$ is projected onto the respective feature map to sample a feature vector $\boldsymbol{\omega}^{(i)}$ trough bilinear interpolation:
\begin{equation}
    \boldsymbol{\omega}^{(i)} = \boldsymbol{W}^{(i)}\left(\boldsymbol{K}^{(i)} \circ \boldsymbol{x}^{(i)}\right).
\label{eq:feat_sampling}
\end{equation}
The NeRF MLP $f$ is split into two parts, $f_1$ and $f_2$. $f_1$ processes the input coordinates of the sampling point alongside the sampled feature vectors into intermediate feature vectors $\boldsymbol{V}^{(i)}$ for every view independently:
\begin{equation}
    \boldsymbol{V}^{(i)} = f_1\left(\boldsymbol{x}^{(i)}, \boldsymbol{d}^{(i)}, \boldsymbol{\omega}^{(i)}\right).
\label{eq:pixelnerf_1}
\end{equation}
The feature vectors from different views are aggregated through average pooling and then processed by $f_2$ to regress the final color and density values:
\begin{equation}
    \boldsymbol{c}, \sigma = f_2\left(\text{mean}_i\left\{\boldsymbol{V}^{(i)}\right\}\right).
\label{eq:pixelnerf_2}
\end{equation}
During training, the $l_1$ distance between estimated ray colors and ground truth RGB values is minimized.
\section{Method}
\label{sec:method}
Given a sparse set of input images \mbox{\small$\{\boldsymbol{I}^{(i)}\}$ ($i=1...N=4$)}, our approach infers a NeRF which allows rendering novel views of the scene. 
We estimate the depth in the given input views and propose two novel techniques to leverage this information during scene reconstruction: (i) the NeRF MLP is conditioned on the difference between sample location and estimated depth which serves as a strong prior for the visual opacity (\Cref{sec:depth_awareness}), and (ii) we focus the scene sampling on the estimated surface regions, i.e. on regions that actually contribute to the scene appearance (\Cref{sec:depth_guided_sampling}). 
Furthermore, we propose to pad and positionally encode the source images prior to the feature extraction to improve extrapolation capabilities beyond the source view frustums (\Cref{sec:feature_extrapolation}).
Please refer to \Cref{fig:method} for an overview of our pipeline.
%


\subsection{Depth Conditioning}
\label{sec:depth_awareness}

Our method is based on pixelNeRF (see \Cref{sec:background}) and leverages the attention-based TransMVSNet~\cite{ding2022transmvsnet} architecture to estimate depth from the sparse observations.
TransMVSNet takes all four input images $\boldsymbol{I}^{(i)}$ as input and predicts the per-view depth maps $\boldsymbol{D}^{(i)}$ as well as the corresponding standard deviations $\boldsymbol{D_\text{std}}^{(i)}$.
%
%
%
%
For each sampling point $\boldsymbol{x}$, we calculate the difference $\Delta_z^{(i)}$ between its z-component in the $i$-th camera space and its corresponding projected depth value:
$\Delta_z^{(i)} = \boldsymbol{D}^{(i)}\left(\boldsymbol{K}^{(i)}\circ\boldsymbol{x}^{(i)}\right) - \boldsymbol{x}^{(i)}_{[z]}$,
which is input to the pixelNeRF MLP $f_1$ as additional conditioning:
\begin{align}
    \boldsymbol{V}^{(i)} &= f_1\left(\boldsymbol{x}^{(i)}, \boldsymbol{d}^{(i)}, \boldsymbol{\omega}^{(i)}, \gamma\left(\Delta_z^{(i)}\right)\right).
    \label{eq:depthawareness}
\end{align}
$\gamma(\cdot)$ denotes positional encoding with exponential frequencies as proposed in \cite{mildenhall2020nerf}.

\subsection{Source Feature Extrapolation}
\label{sec:feature_extrapolation}
%
%
When projecting sampling points on the source feature maps, image-based NeRFs typically apply border padding, i.e. points outside the map's boundaries are assigned constant feature vectors irrespective of their distance to the feature map. 
During synthesis, this causes smearing artifacts in regions that are not visible by the source images (see \Cref{fig:ablation}, third column).
To solve this, we make two modifications to the source images $\boldsymbol{I}^{(i)}$ \emph{before} applying the encoder network $\mathcal{E}$.
We apply border padding and add channels with positional encodings of the padded pixel coordinates, resulting in $\boldsymbol{I'}^{(i)} = concatenate\left(\boldsymbol{I}^{(i)}_\text{pad}, \boldsymbol{\Gamma}\right)$,
where $\boldsymbol{\Gamma}$ contains the pixel-wise positional encodings for the padded regions:
\begin{align}
    \begin{split}
        \boldsymbol{\Gamma}_{[u, v]} &= 
    \begin{cases}
           \gamma(u,v)&\text{if } (u, v) \not\in \boldsymbol{I}^{(i)} \\
        \hfill 0 \hfill &\text{if } (u, v) \in \boldsymbol{I}^{(i)} .
    \end{cases}
    \end{split}
\end{align}
%
%
%
The positional encoding supports $\mathcal{E}$ in regressing distinctive features in padded regions where the extrapolated color values are constant. 
%
%

\subsection{Depth-Guided Sampling}
\label{sec:depth_guided_sampling}

\begin{figure*}[t]
    \centering
    \vspace{-0.25cm}
    \includegraphics[width=.9\textwidth]{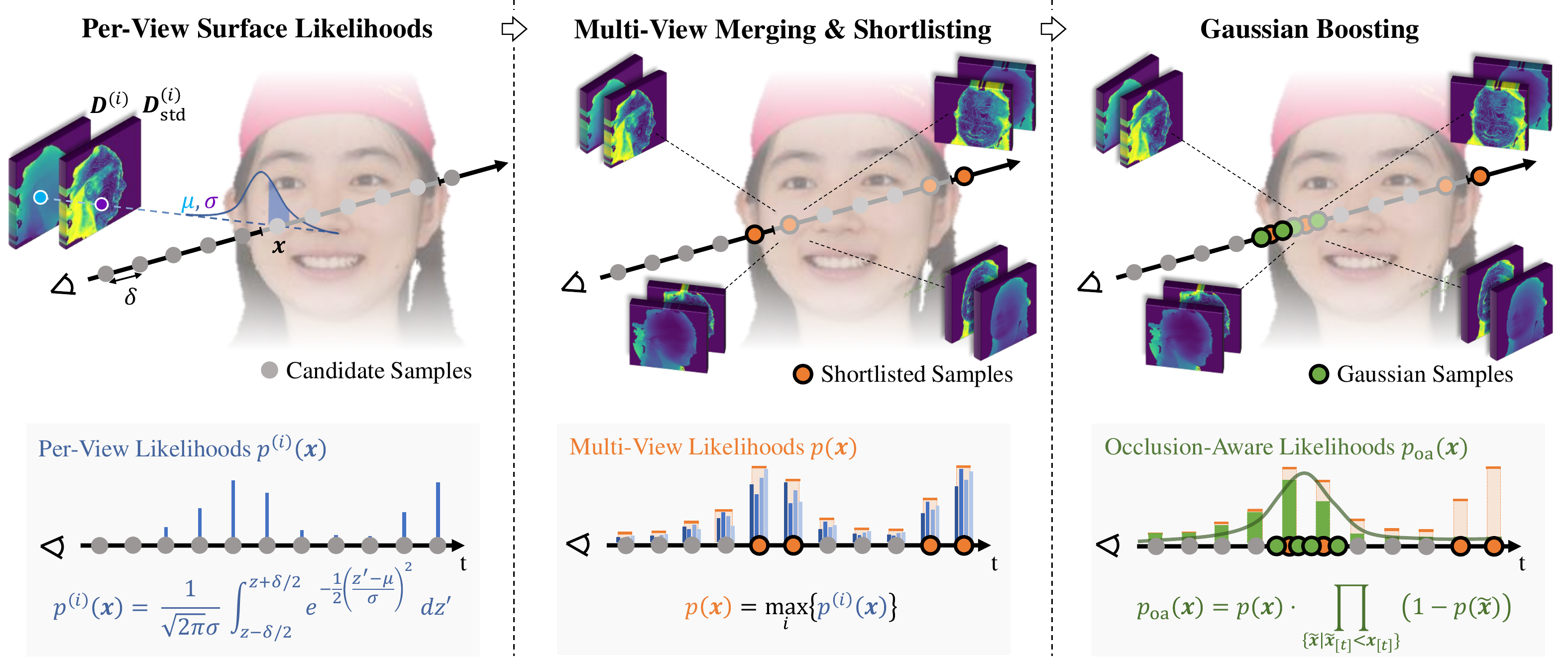}
    \vspace{-0.25cm}
    \caption{Depth-guided sampling. We sample candidate points along the target ray and evaluate their surface likelihoods given the depth estimates for each input view. The view-wise likelihoods are aggregated through max-pooling and we shortlist the most-likely samples. Additional points are sampled according to a Gaussian distribution that was fitted against the occlusion-aware likelihoods of all candidates.}
    \vspace{-0.25cm}
    \label{fig:depth_guided_sampling}
\end{figure*}
Since only object surfaces and the immediate surroundings contribute to ray colors, we aim to focus our sampling on these regions. 
The estimated depth maps provide strong priors about the placements of such surfaces. This allows us to increase the sample density in relevant regions which improves the synthesis quality (see \Cref{sec:ablation}). \Cref{fig:depth_guided_sampling} provides an overview of our approach.
Note that while previous work incorporates depth information of target views during NeRF training \cite{rematas2021urban, roessle2022depthpriorsnerf, kangle2021dsnerf}, we are the first to exploit depth from input views that do not coincide with the target view and which we predict both at training and test time.
%

\vspace{-0.35cm}
\paragraph{Depth-Guided Probability Fields}
For each input image, the depth estimator provides pixel-wise depth expectation values $\boldsymbol{D}^{(i)}$ and standard deviations $\boldsymbol{D_\text{std}}^{(i)}$. 
These maps define pixel-aligned probability density fields for the presence of object surfaces.  
Assuming a ray \mbox{$\boldsymbol{r}(t)$} with near plane $t_\text{near}$ and far plane $t_\text{far}$, we first uniformly sample a large set of $N_\text{cand}$ candidate samples along the ray :
\begin{equation}
    \{\boldsymbol{x}\} = \left\{\boldsymbol{r}(t) ~|~t\sim\left[t_\text{near}, t_\text{far}\right] \right\} .
\end{equation}
For each of the input views, we project $\{\boldsymbol{x}\}$ onto the respective depth maps and determine the likelihood of $\boldsymbol{x}$ being a surface point:
\vspace{-0.1cm}
\begin{equation}
    p^{(i)}(\boldsymbol{x}) = \frac{1}{\sqrt{2\pi}\sigma} \int_{z-\delta/2}^{z+\delta/2} e^{-\frac{1}{2}\left(\frac{z'-\mu}{\sigma}\right)^2} dz' ,
\end{equation}
\vspace{-0.2cm}
{\small
\begin{align*}
\text{with}\quad \mu &= \boldsymbol{D}^{(i)}\left(\boldsymbol{K}^{(i)}\circ\boldsymbol{x}^{(i)}\right), & z &= \boldsymbol{x}^{(i)}_{[z]}, \\
\sigma &= \boldsymbol{D}_\text{std}^{(i)}\left(\boldsymbol{K}^{(i)}\circ\boldsymbol{x}^{(i)}\right),&\delta &= (t_\text{far}-t_\text{near})/N_\text{cand}.
\end{align*}
}%


\noindent
We perform backface culling by first calculating normals from each depth map and then discarding samples if the angle between the ray direction and the projected normal value is smaller than $90^\circ$.  
The likelihood $p(\boldsymbol{x})$ of a point coinciding with a surface is determined by view-wise max-pooling:
\begin{equation}
    p(\boldsymbol{x}) = \max_i\left\{p^{(i)}(\boldsymbol{x})\right\}.
\end{equation}
We shortlist $N_\text{samples}$ candidates with the highest likelihoods for sampling the NeRF. 
%

\vspace{-0.35cm}
\paragraph{Gaussian Boosting}
To further improve sampling efficiency, we sample additional points around the termination expectation value of the ray. 
%
The occlusion-aware likelihoods of the ray $\boldsymbol{r}$ terminating in sample $\boldsymbol{x}$ is given by:
\begin{equation}
    p_\text{oa}(\boldsymbol{x}) = p(\boldsymbol{x}) \cdot \prod_{\left\{\boldsymbol{\tilde{x}}~|~\boldsymbol{\tilde{x}}_{[t]}<\boldsymbol{x}_{[t]}\right\}} \left(1-p(\boldsymbol{\tilde{x}}\right)). 
\end{equation}
Please note the simplified notation \mbox{$\boldsymbol{x}_{[t]} \coloneqq t$} so that \mbox{$\boldsymbol{r}(t) = \boldsymbol{x}$}.
We fit a Gaussian distribution against the occlusion-aware likelihoods along the ray (see \Cref{fig:depth_guided_sampling} right) and sample $N_\text{gauss}$ points from it which are added to the shortlisted candidates. 
%

%
%
%

\subsection{Loss Formulation}
\label{sec:perc_loss}
\begin{figure*}[ht!]
    \centering
    \def\svgwidth{.9\linewidth}
    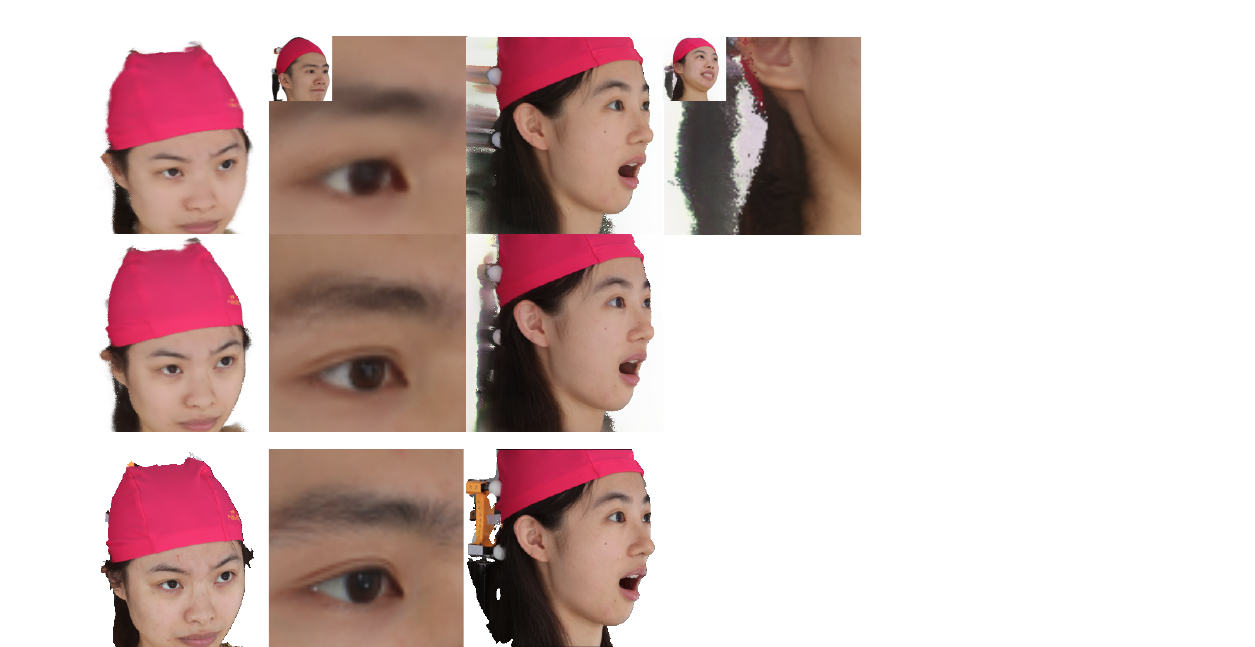
    \caption{Qualitative ablation study. Starting with pixelNeRF~\cite{yu2020pixelnerf} (first row, first column) as a baseline, we progressively add the components of our method and demonstrate their effects until we reach our final model (second row, last column).}
    \label{fig:ablation}
\end{figure*}

Our loss formulation consists of a per-pixel reconstruction error $\mathcal{L}_{l_1}$ as well as a perceptual loss $\mathcal{L}_\text{vgg}$~\cite{vggloss}.
While a perceptual loss improves high-frequency details, it can introduce color shifts (see \Cref{fig:ablation}). 
Thus, we introduce an anti-bias term $\mathcal{L}_\text{ab}$ which corresponds to a standard $l_1$ loss that is applied to downsampled versions of prediction and ground truth.
$\mathcal{L}_\text{ab}$ effectively eliminates color shifts while being robust against minor misalignments, hence in contrast to the standard $l_1$ loss, it does not introduce low-frequency bias. 
Let $\boldsymbol{P}$ denote the ground truth image patch and $\boldsymbol{\hat{P}}$ its predicted counterpart, then $\mathcal{L}_\text{ab}$ is given by:
\begin{equation}
    \mathcal{L}_\text{ab} = \left|\left|\text{DS}_k(\boldsymbol{P}) - \text{DS}_k(\boldsymbol{\hat{P}})\right|\right|_1^1 ,
\end{equation}
where $\text{DS}_k(\cdot)$ denotes $k$-fold downsampling.
Our full objective function is defined as:
\begin{equation}    
    \mathcal{L} = w_{l_1} \cdot \mathcal{L}_{l_1} + w_\text{vgg} \cdot \mathcal{L}_\text{vgg} + w_\text{ab} \cdot \mathcal{L}_\text{ab} ,
\end{equation}
where $\mathcal{L}_{l_1}$ corresponds to a pixel-wise $l_1$ distance.
%
%
%
We use an Adam optimizer with standard parameters and a learning rate of $10^{-4}$ and train on patches of $64\times64$ px on a single NVIDIA A100-SXM-80GB GPU with batch size 4 for 330k iterations which takes 4 days. 
Note that depth-guided sampling allows reducing the samples per ray by a factor of 4 w.r.t. pixelNeRF because samples are focused on non-empty areas which improves the performance (see \Cref{tab:ablation}).
%

\begin{figure*}[ht!]
    \vspace{0.4cm}
    \centering
    \def\svgwidth{.95\linewidth}
    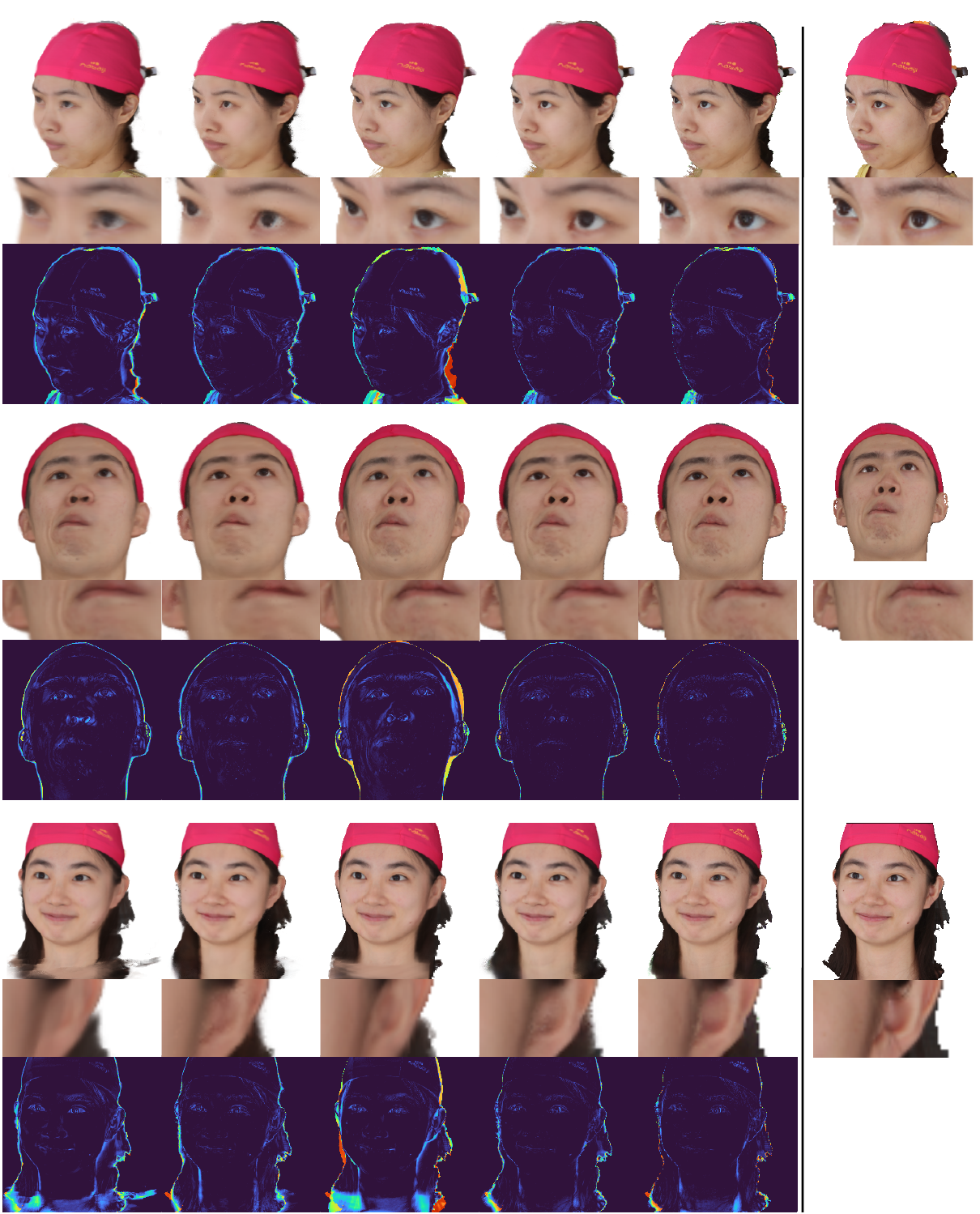
    \caption{Qualitative comparison on FaceScape. IBRNet and KeypointNeRF produce artifacts for regions outside of the source images (\textcolor{mypink}{pink}). pixelNeRF handles these aspects better but produces blurry results. KeypointNeRF synthesizes heads with deformations (\textcolor{myorange}{orange}). Even without the perceptual loss, our method yields better results and adding the perceptual loss further emphasizes high-frequency details.
    }
    \label{fig:qual_comparison_facescape}
\end{figure*}

\section{Results}
\label{sec:results}
We conduct the validation of our approach on the FaceScape dataset~\cite{yang2020facescape} which contains more than 400k portrait photographs of 359 subjects under 20 different expressions captured with a rig of approximately 60 cameras, providing camera parameters and a 3D mesh that can be used to render the ground truth depth maps and segmentations for training. 
%
Four of the captured subjects consented to the public display of their data which determined our training and validation splits.  
We apply post-processing in terms of color calibration and unification of depth ranges, crop the images around the face region and rescale it to $256\times256$ px.
%
%
We sample source view quadruples spanning horizontal and vertical angles of 45° and 30° respectively. Target views are sampled between the source views which results in $500k$ samples for training and $7k$ for validation. 
%

%

%

%
\vspace{-0.35cm}
\paragraph{Baselines}
We compare our approach against the state-of-the-art methods pixelNeRF~\cite{yu2020pixelnerf}, IBRNet~\cite{wang2021ibrnet}, and KeypointNeRF~\cite{Mihajlovic:ECCV2022} using the author's codebases.
pixelNeRF~\cite{yu2020pixelnerf} enables NeRFs to generalize across scenes by conditioning them on feature maps extracted from posed source views.
IBRNet~\cite{wang2021ibrnet} adds a transformer architecture and estimates blend weights for the source images instead of regressing color directly.
KeypointNeRF~\cite{Mihajlovic:ECCV2022} adds triangulated landmarks to guide the geometry estimation. 
In the official implementation, KeypointNeRF only renders the intersection of the source view frustums which we changed to their union in order to be able to cope with partial observations in source views. 
Since off-the-shelf keypoint detectors struggle with strong head rotations, we provide KeypointNeRF with ground truth keypoints.
We also evaluate a version of our method that does not include the perceptual loss and anti-bias term (see \Cref{sec:perc_loss}).

\vspace{-0.35cm}
\paragraph{Metrics}
We quantitatively evaluate the performance of our model through the pixel-wise Manhattan and Euclidean distances (L1) and (L2), structural similarity (SSIM), learned perceptual image patch similarity (LPIPS), and the peak signal-to-noise ratio (PSNR).


\subsection{Novel View Synthesis}
\begin{table}
    \resizebox{.9\linewidth}{!}{%
    \begin{tabular}{l|ccccc} \toprule
    \textbf{Method}& \textbf{LPIPS} $\downarrow$& \textbf{PSNR} $\uparrow$& \textbf{SSIM} $\uparrow$& \textbf{L1} $\downarrow$& \textbf{L2} $\downarrow$\\ \midrule
    IBRNet \cite{wang2021ibrnet}& $0.159$& $22.7$& $0.89$& $0.025$& $0.006$\\
    pixelNeRF \cite{yu2020pixelnerf}& $0.165$& $23.54$& $0.90$ & $0.021$& $0.005$\\
    KeypointNeRF \cite{Mihajlovic:ECCV2022}& $0.148$& $18.39$& $0.86$& $0.036$& $0.017$\\\midrule
    Ours w/o $\mathcal{L}_\text{vgg}$& $0.137$& $\mathbf{24.40}$& $\mathbf{0.92}$ & $\mathbf{0.018}$& $\mathbf{0.004}$\\
    Ours& $\mathbf{0.099}$& $22.42$& $0.91$ & $0.020$& $0.007$\\\bottomrule
    \end{tabular}
    }
    \caption{Quantitative comparisons on FaceScape show that our method has a significantly lower perceptual error in comparison to state-of-the-art methods while having on-par pixel-wise errors.}
    \label{tab:quant_comparison_facescape}
\end{table}

\Cref{fig:qual_comparison_facescape} displays a qualitative comparison between our method and the baselines on the FaceScape dataset~\cite{yang2020facescape}.
IBRNet~\cite{wang2021ibrnet} generally performs well in regions that project onto large areas on the source images, e.g., wrinkles in the mouth region for extreme expressions are synthesized convincingly. However, especially the eye regions project onto small areas and IBRNet fails to produce plausible results. Regions that lie outside of the source views show significant artifacts. 
pixelNeRF~\cite{yu2020pixelnerf} solves these cases better but tends to produce blurry results. 
KeypointNeRF~\cite{Mihajlovic:ECCV2022} synthesizes high-frequency details very plausibly but shows artifacts for regions outside of the source views and we also noticed severe deformations of the overall head shape. 
We attribute these deformations to the sparsity of the triangulated landmarks.
The dense guidance by depth maps in our method effectively solves this artifact and similarly to pixelNeRF, regressing color values directly allows us to plausibly synthesize regions that lie outside of the source views.
At the same time, even without a perceptual loss, our method synthesizes high-frequency details better than IBRNet and pixelNeRF.
Adding a perceptual loss emphasizes high-frequency detail synthesis even more and yields results that qualitatively outperform all baselines even though pixel-wise scores slightly worsen (see \Cref{tab:quant_comparison_facescape}).


\subsection{Ablation Study}
\label{sec:ablation}
Since our approach is based on pixelNeRF~\cite{yu2020pixelnerf}, we perform an additive ablation study to evaluate the contributions of our changes in which we progressively add our novel components and discuss their effects on the model performance. 
%
\Cref{fig:ablation} visualizes the qualitative effects by progressively introducing one new component per column. \Cref{tab:ablation} provides the quantitative evaluation. 
First, we add depth awareness to pixelNeRF (\Cref{sec:depth_awareness}) which improves the overall synthesis quality and all scores. 
The perceptual loss $\mathcal{L}_\text{vgg}$ adds more high-frequency details but pixel-wise scores degrade slightly.
%
Source feature extrapolation (\Cref{sec:feature_extrapolation}) counters smearing artifacts that occur in regions that are invisible in the source views. 
Color shifts that appeared after adding $\mathcal{L}_\text{vgg}$ can be eliminated with the anti-bias loss $\mathcal{L}_\text{ab}$ (\Cref{sec:perc_loss}). 
Depth-guided sampling (\Cref{sec:depth_guided_sampling}) increases the sampling density around the head surface and especially improves the synthesis quality of thin surfaces like ears. 
As a side effect, it also allows us to reduce the number of samples per ray and increase the batch size from 1 to 4 without changing GPU memory requirements. This stabilizes training such that minor artifacts vanish and consistently improves all metrics. 
%

\begin{table}
    \resizebox{\linewidth}{!}{%
\begin{tabular}{l|ccccc} \toprule
\textbf{Method}& \textbf{LPIPS} $\downarrow$& \textbf{L1} $\downarrow$& \textbf{L2} $\downarrow$& \textbf{PSNR} $\uparrow$& \textbf{SSIM} $\uparrow$\\ \midrule
pixelNeRF \cite{yu2020pixelnerf} & $0.16$& $0.021$& $\mathbf{0.005}$& $23.54$& $0.90$\\
+ Depth Awareness & $0.15$ & $\mathbf{0.020}$ & $\mathbf{0.005}$ & $\mathbf{23.71}$ & $\mathbf{0.91}$ \\
+ Perc. Loss $\mathcal{L}_\text{vgg}$& $0.11$ & $0.032$ & $0.008$ &  $21.66$ & $0.89$ \\
+ Source Feature Extrapolation  & $0.11$ & $0.029$ & $0.008$ &  $21.96$ & $0.90$ \\
+ Anti-bias Loss $\mathcal{L}_\text{ab}$ & $0.11$ & $0.022$ & $0.008$ & $22.02$ & $0.90$ \\
+ Depth-Guided Sampling  & $0.12$ & $0.023$ & $0.008$ & $21.95$ & $0.90$ \\
+ Increased Batch Size & $\mathbf{0.10}$ & $\mathbf{0.020}$ & $0.007$ &  $22.42$ & $\mathbf{0.91}$  \\
\bottomrule
\end{tabular}
}
\caption{Quantitative ablation study on FaceScape~\cite{yang2020facescape}.}
\vspace{-0.4cm}
\label{tab:ablation}
\end{table}


\subsection{Generalization to out-of-train-distribution data}
\begin{figure}[ht!]
    \centering
    \def\svgwidth{\linewidth}
\begingroup%
  \makeatletter%
  \providecommand\color[2][]{%
    \errmessage{(Inkscape) Color is used for the text in Inkscape, but the package 'color.sty' is not loaded}%
    \renewcommand\color[2][]{}%
  }%
  \providecommand\transparent[1]{%
    \errmessage{(Inkscape) Transparency is used (non-zero) for the text in Inkscape, but the package 'transparent.sty' is not loaded}%
    \renewcommand\transparent[1]{}%
  }%
  \providecommand\rotatebox[2]{#2}%
  \newcommand*\fsize{\dimexpr\f@size pt\relax}%
  \newcommand*\lineheight[1]{\fontsize{\fsize}{#1\fsize}\selectfont}%
  \ifx\svgwidth\undefined%
    \setlength{\unitlength}{364.47459303bp}%
    \ifx\svgscale\undefined%
      \relax%
    \else%
      \setlength{\unitlength}{\unitlength * \real{\svgscale}}%
    \fi%
  \else%
    \setlength{\unitlength}{\svgwidth}%
  \fi%
  \global\let\svgwidth\undefined%
  \global\let\svgscale\undefined%
  \makeatother%
  \begin{picture}(1,0.45735077)%
    \lineheight{1}%
    \setlength\tabcolsep{0pt}%
    \scriptsize
    \put(0,0){\includegraphics[width=\unitlength,page=1]{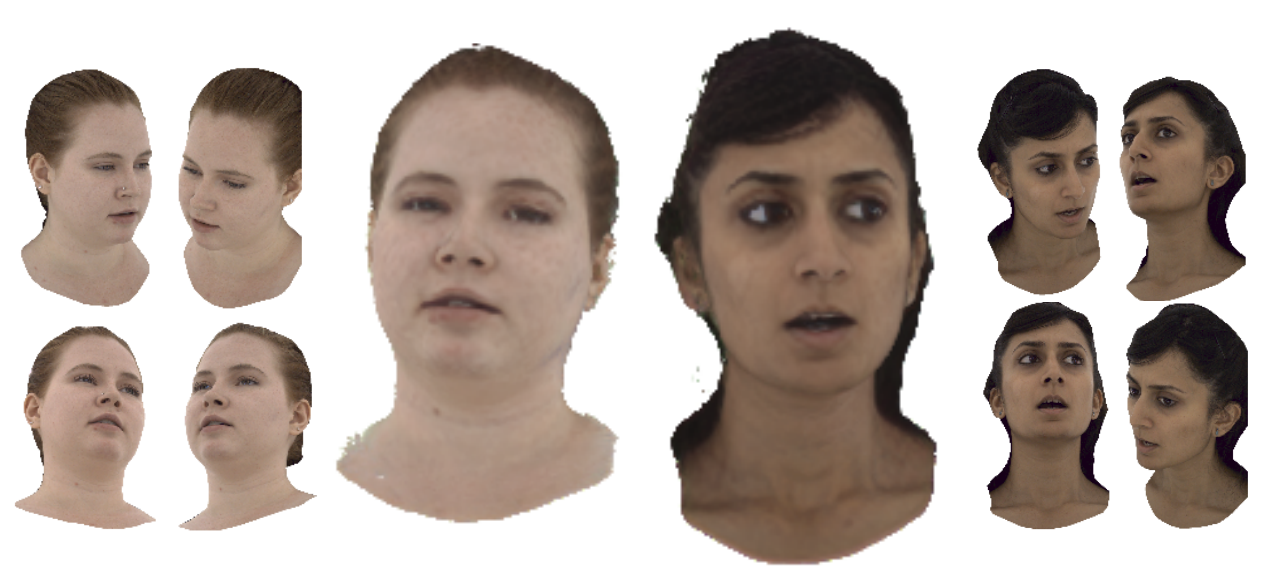}}%
    \put(0.13253656,0.44484344){\makebox(0,0)[t]{\lineheight{1.25}\smash{\begin{tabular}[t]{c}Source Views\end{tabular}}}}%
    \put(0.38568395,0.44484344){\makebox(0,0)[t]{\lineheight{1.25}\smash{\begin{tabular}[t]{c}Prediction\end{tabular}}}}%
    \put(0.63247123,0.44484344){\makebox(0,0)[t]{\lineheight{1.25}\smash{\begin{tabular}[t]{c}Prediction\end{tabular}}}}%
    \put(0.87724627,0.44484344){\makebox(0,0)[t]{\lineheight{1.25}\smash{\begin{tabular}[t]{c}Source Views\end{tabular}}}}%
  \end{picture}%
\endgroup%

    \caption{Generalization to out-of-train-distribution samples with different ethnicities and headwear.}
    \vspace{-0.35cm}
    \label{fig:multiface}
\end{figure}

The Facescape dataset \cite{yang2020facescape} used for training DINER is restricted to subjects of Asian ethnicity wearing a red swimming cap. 
However, especially for video-conferencing applications it is crucial that models generalize well to out-of-train-distribution data to prevent discrimination of underrepresented minorities. 
We demonstrate the generalization capabilities of DINER in \cref{fig:multiface} by applying it to two subjects from the Multiface dataset~\cite{wuu2022multiface}. Despite their highly different ethnicities and headwear, DINER yields plausible synthesis results. In the suppl. video we further show that despite only being trained on static frames, DINER exhibits high temporal stability for continuous sequences.

\subsection{Novel View Synthesis on General Objects}

\begin{figure}[t]
    \centering
    \def\svgwidth{\linewidth}
\begingroup%
  \makeatletter%
  \providecommand\color[2][]{%
    \errmessage{(Inkscape) Color is used for the text in Inkscape, but the package 'color.sty' is not loaded}%
    \renewcommand\color[2][]{}%
  }%
  \providecommand\transparent[1]{%
    \errmessage{(Inkscape) Transparency is used (non-zero) for the text in Inkscape, but the package 'transparent.sty' is not loaded}%
    \renewcommand\transparent[1]{}%
  }%
  \providecommand\rotatebox[2]{#2}%
  \newcommand*\fsize{\dimexpr\f@size pt\relax}%
  \newcommand*\lineheight[1]{\fontsize{\fsize}{#1\fsize}\selectfont}%
  \ifx\svgwidth\undefined%
    \setlength{\unitlength}{230.05459763bp}%
    \ifx\svgscale\undefined%
      \relax%
    \else%
      \setlength{\unitlength}{\unitlength * \real{\svgscale}}%
    \fi%
  \else%
    \setlength{\unitlength}{\svgwidth}%
  \fi%
  \global\let\svgwidth\undefined%
  \global\let\svgscale\undefined%
  \makeatother%
  \begin{picture}(1,0.44667758)%
    \lineheight{1}%
    \scriptsize
    \setlength\tabcolsep{0pt}%
    \put(0,0){\includegraphics[width=\unitlength,page=1]{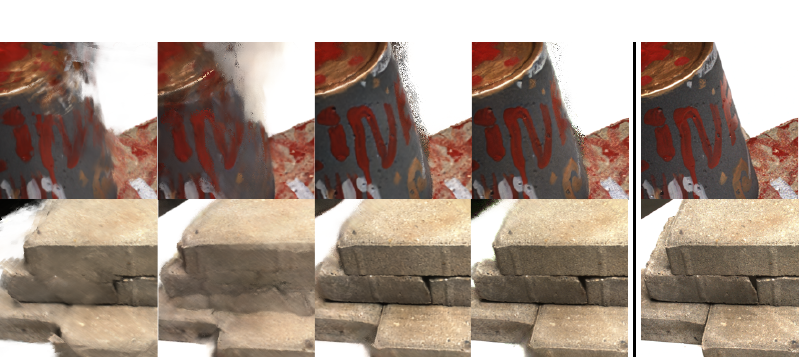}}%
    \put(0.09713244,0.4235598){\color[rgb]{0,0,0}\makebox(0,0)[t]{\lineheight{1.25}\smash{\begin{tabular}[t]{c}IBRNet \cite{wang2021ibrnet}\end{tabular}}}}%
    \put(0.29645379,0.4235598){\color[rgb]{0,0,0}\makebox(0,0)[t]{\lineheight{1.25}\smash{\begin{tabular}[t]{c}pixelNeRF \cite{yu2020pixelnerf}\end{tabular}}}}%
    \put(0.4924631,0.4235598){\color[rgb]{0,0,0}\makebox(0,0)[t]{\lineheight{1.25}\smash{\begin{tabular}[t]{c}Ours w/o $\mathcal{L}_\text{perc}$\end{tabular}}}}%
    \put(0.68878573,0.4235598){\color[rgb]{0,0,0}\makebox(0,0)[t]{\lineheight{1.25}\smash{\begin{tabular}[t]{c}Ours\end{tabular}}}}%
    \put(0.90141285,0.4235598){\color[rgb]{0,0,0}\makebox(0,0)[t]{\lineheight{1.25}\smash{\begin{tabular}[t]{c}Ground Truth\end{tabular}}}}%
    \put(0,0){\includegraphics[width=\unitlength,page=2]{qual_eval_dtu.pdf}}%
  \end{picture}%
\endgroup%

    \caption{Qualitative comparison on DTU~\cite{jensen2014large} which consists of a variety of objects like a bucket (1rst row) or bricks (2nd row).}
    \vspace{-0.35cm}
    \label{fig:qual_comparison_dtu}
\end{figure}

We further evaluate DINER on general objects in the DTU dataset~\cite{jensen2014large}. 
%
%
We follow the training/validation split convention proposed in \cite{yu2020pixelnerf}, adopt the preprocessing steps from \cite{ding2022transmvsnet}, and apply additional 2-fold downsampling to obtain images with resolution $256\times320$ px. 
Similar to the training on FaceScape, we sample source view quadruples spanning horizontal and vertical angles of $\approx 50^\circ$ and $\approx 35^\circ$ respectively which results in $30k$ samples for training and $5k$ for validation.
In \Cref{fig:qual_comparison_dtu}, we show the comparison of our method to the baselines.
Note that KeypointNeRF is class-specific and can not be applied to general objects.
The results demonstrate that our method outperforms all baselines by a significant margin, see \Cref{tab:quant_comparison_dtu} (suppl. material) for the quantitative evaluation. 
%


%
\section{Discussion}
%
DINER excels at synthesizing photo-realistic 3D scenes from few input images. Still, some challenges remain before it may be used for real-world applications such as immersive video conferencing. 
Similar to most NeRF-based methods, our rendering speed is slow.
Despite improved sampling efficiency through depth guidance, the synthesis of a $256^2$ px image still takes two seconds. 
While real-time-capable NeRF methods exist \cite{mueller2022instant, yu2021plenoctrees, fastnerf}, none of them generalizes across scenes yet.
%
%
As our method relies on a depth prediction network, we are bound by its accuracy.
However, it could be replaced by Kinect-like depth sensors.
%

%
\section{Conclusion}
We presented depth-aware image-based neural radiance fields (DINER) which synthesize photo-realistic 3D scenes given only four input images. 
To capture scenes more completely and with high visual quality, we assume to have input images with a high disparity.
We leverage a state-of-the-art depth estimator to guide the implicit geometry estimation and to improve sampling efficiency. 
In addition, we propose a technique to extrapolate features from the input images.
Our experiments show that DINER outperforms the state of the art both qualitatively and quantitatively. 
DINER's ability to reconstruct both human heads as well as general objects with high quality is vital for real-world applications like immersive video conferencing with holographic displays.
%
%

\section{Acknowledgements}
This project has received funding from the Max Planck ETH Center for Learning Systems (CLS). Further, we would like to thank Marcel C. Bühler, Philip-William Grassal, Yufeng Zheng, Lixin Xue, and Xu Chen for their valuable feedback. 

\clearpage
\appendix

\twocolumn[{%
\renewcommand\twocolumn[1][]{#1}%
\maketitle
\begin{center}
    \begin{spacing}{2}\textbf{\Large{DINER: Depth-aware Image-based NEural Radiance fields \\ -- Supplemental Document --}}
    \end{spacing}
    \vspace{1cm}
    \centering
    \captionsetup{type=figure}
    \def\svgwidth{.85\linewidth}
\begingroup%
  \makeatletter%
  \providecommand\color[2][]{%
    \errmessage{(Inkscape) Color is used for the text in Inkscape, but the package 'color.sty' is not loaded}%
    \renewcommand\color[2][]{}%
  }%
  \providecommand\transparent[1]{%
    \errmessage{(Inkscape) Transparency is used (non-zero) for the text in Inkscape, but the package 'transparent.sty' is not loaded}%
    \renewcommand\transparent[1]{}%
  }%
  \providecommand\rotatebox[2]{#2}%
  \newcommand*\fsize{\dimexpr\f@size pt\relax}%
  \newcommand*\lineheight[1]{\fontsize{\fsize}{#1\fsize}\selectfont}%
  \ifx\svgwidth\undefined%
    \setlength{\unitlength}{595.27559055bp}%
    \ifx\svgscale\undefined%
      \relax%
    \else%
      \setlength{\unitlength}{\unitlength * \real{\svgscale}}%
    \fi%
  \else%
    \setlength{\unitlength}{\svgwidth}%
  \fi%
  \global\let\svgwidth\undefined%
  \global\let\svgscale\undefined%
  \makeatother%
  \begin{picture}(1,0.50581977)%
    \lineheight{1}%
    \setlength\tabcolsep{0pt}%
    \put(0,0){\includegraphics[width=\unitlength,page=1]{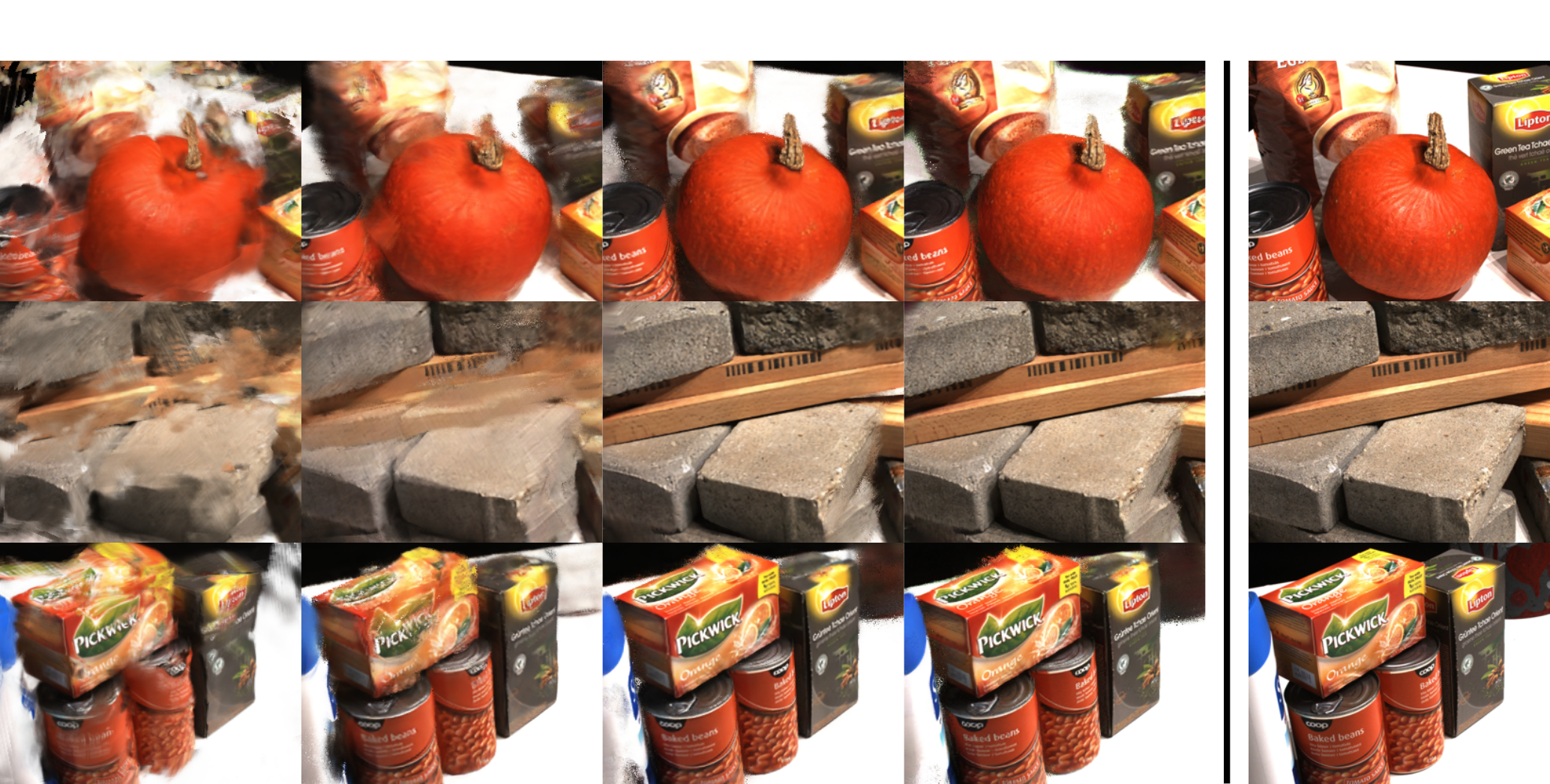}}%
    \put(0.09633255,0.48795121){\makebox(0,0)[t]{\lineheight{1.25}\smash{\begin{tabular}[t]{c}IBRNet \cite{wang2021ibrnet}\end{tabular}}}}%
    \put(0.29074317,0.48795121){\makebox(0,0)[t]{\lineheight{1.25}\smash{\begin{tabular}[t]{c}PixelNeRF \cite{yu2020pixelnerf}\end{tabular}}}}%
    \put(0.48515378,0.48795121){\makebox(0,0)[t]{\lineheight{1.25}\smash{\begin{tabular}[t]{c}Ours w/o $\mathcal{L}_\text{vgg}$\end{tabular}}}}%
    \put(0.67956441,0.48795121){\makebox(0,0)[t]{\lineheight{1.25}\smash{\begin{tabular}[t]{c}Ours\end{tabular}}}}%
    \put(0.90192188,0.48795121){\makebox(0,0)[t]{\lineheight{1.25}\smash{\begin{tabular}[t]{c}Ground Truth\end{tabular}}}}%
  \end{picture}%
\endgroup%

      \caption{
        Qualitative comparison on general objects of the DTU dataset~\cite{jensen2014large}.
        Our depth-aware image-based neural radiance field shows significantly higher image quality with fewer distortions and blurring artifacts.
      }
      \vspace{0.1cm}
      \label{fig:dtu_qual_2}
\end{center}%
}]

\begin{abstract}
    In this supplemental document, we detail the architecture of our method \emph{DINER} (see \Cref{sec:arch}), provide a quantitative comparison to state-of-the-art models on novel view synthesis for general objects in the DTU dataset (see \Cref{sec:dtu}), evaluate the influence of the depth estimator's accuracy on the synthesis quality (see \Cref{sec:depth_acc}), and conduct further experiments concerning depth-guided sampling (see \Cref{sec:eval_depth_guided_sampling}). We conclude this document with a discussion of ethical implications of our work (see \Cref{sec:ethical_considerations}).
\end{abstract}

\begin{figure*}[t]
    \centering
    \begin{subfigure}[b]{0.33\textwidth}
        \centering
        \raisebox{-0.75 cm}{\input{figs/DepthNoiseAblation.pgf}}
    \end{subfigure}
    \hfill
    \begin{subfigure}[b]{0.65\textwidth}
        \def\svgwidth{\linewidth}
\begingroup%
  \makeatletter%
  \providecommand\color[2][]{%
    \errmessage{(Inkscape) Color is used for the text in Inkscape, but the package 'color.sty' is not loaded}%
    \renewcommand\color[2][]{}%
  }%
  \providecommand\transparent[1]{%
    \errmessage{(Inkscape) Transparency is used (non-zero) for the text in Inkscape, but the package 'transparent.sty' is not loaded}%
    \renewcommand\transparent[1]{}%
  }%
  \providecommand\rotatebox[2]{#2}%
  \newcommand*\fsize{\dimexpr\f@size pt\relax}%
  \newcommand*\lineheight[1]{\fontsize{\fsize}{#1\fsize}\selectfont}%
  \ifx\svgwidth\undefined%
    \setlength{\unitlength}{226.93424255bp}%
    \ifx\svgscale\undefined%
      \relax%
    \else%
      \setlength{\unitlength}{\unitlength * \real{\svgscale}}%
    \fi%
  \else%
    \setlength{\unitlength}{\svgwidth}%
  \fi%
  \global\let\svgwidth\undefined%
  \global\let\svgscale\undefined%
  \makeatother%
  \begin{picture}(1,0.23134441)%
    \lineheight{1}%
    \setlength\tabcolsep{0pt}%
    \put(-0.00680972,0.020){\color[rgb]{0,0,0}\makebox(0,0)[lt]{\begin{minipage}{\unitlength}\scriptsize\begin{tabularx}{\textwidth}{*5{>{\centering\arraybackslash}X}} 10 & 6.0 & 1.0 & 0.6 & Ground Truth\\ \end{tabularx}\end{minipage}}}%
    \put(0,0){\includegraphics[width=\unitlength,page=1]{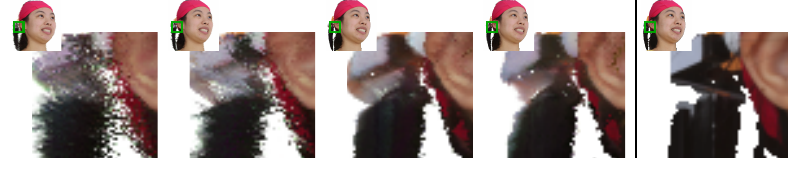}}%
  \end{picture}%
\endgroup%

    \end{subfigure}  
    \vspace{-0.5cm}
    \caption{Model performance under noisy depth signals. The synthesis quality improves with increasing depth accuracy up to a standard deviation of 1mm. Depth information with even higher accuracy does not yield further improvements in terms of synthesis quality.}
    \label{fig:depth_noise_ablation}
\end{figure*}


\renewcommand{\thesection}{\Alph{section}}

\section{Architecture Details}
\label{sec:arch}

We adopt the model architecture of pixelNeRF~\cite{yu2020pixelnerf} and kindly refer to their supplemental material for further details about the image encoder and the NeRF network. 
Our newly introduced components require two adaptations, namely when we introduce depth awareness we change the dimensionality of the feature vector that conditions the MLP, and the source feature extrapolation requires us to change the input channel size of the image encoder.
Both adaptations will be detailed in the following paragraphs.

\paragraph{Depth Awareness}
To guide the scene reconstruction, we also condition the NeRF on the positionally encoded distance between the z-coordinate of the sampling point in camera coordinates and the projected depth value. 
We employ the same positional encoding as in the original NeRF~\cite{mildenhall2020nerf} and use 6 frequency channels with a base frequency of $1\frac{1}{\text{meter}}$.
The resulting 13-dimensional vector is concatenated with the 512-dimensional feature vector sampled from the feature maps and then used to condition the NeRF MLP. The input layer weight dimensions of the MLP are adjusted accordingly. 

\paragraph{Source Feature Extrapolation}
We use a combination of image padding and positional encoding to enable the image encoder to extrapolate the feature maps. 
The images are padded by 64 px by repeating the border values. The positional encoding ranges over 4 exponentially increasing frequencies starting with $0.5$ and is applied to the pixel's uv coordinates which are normalized to $[-1,+1]$. 
The resulting positional encoding map has a channel size of 18. Note that the positional encoding is set to 0 for all pixels that do not belong to the padded region.   
Adding positional encodings to the source image before applying the image encoder means that the inputs to the image encoder no longer have 3 channels.  Since we employ a pretrained network, we have to add randomly initialized weights to its first layer. 
Note that because the positional encoding maps are set to zero in unpadded regions, here the added weights do not have an effect on the predictions of the pretrained network. 

\paragraph{Depth-Guided Sampling}
For depth-guided sampling, we use 1000 candidate samples per ray from which we shortlist 25 samples and add 15 samples during Gaussian boosting. This sums up to 40 samples in total which contribute to the final ray color. The normal maps that we require for point cloud backface culling are obtained by calculating the central difference on the depth maps via convolutional kernels with size 3. Foreground-background edges are filtered out.

\paragraph{Objective Function}
The objective function for training DINER consists of 3 terms: a pixel-wise $l_1$ distance $\mathcal{L}_{l_1}$, a perceptual loss $\mathcal{L}_\text{vgg}$, and the anti-bias term $\mathcal{L}_\text{ab}$. The according weights are
\begin{align*}
    w_{l_1}&=1.0\\
    w_\text{vgg}&=0.1\\
    w_\text{ab}&=5.0 \scriptstyle ~(1.0 ~\text{for DTU}).
\end{align*}
All terms are evaluated on patches of $64\times64$ px unless noted otherwise. $\mathcal{L}_\text{ab}$ downsamples the patches to $8\times8$ px through average pooling before evaluating the $l_1$ distance.
The perceptual loss was adopted from \cite{Mihajlovic:ECCV2022}.


\section{Further Comparisons on DTU}
\label{sec:dtu}
\begin{figure}[h]
    \centering
    \def\svgwidth{\linewidth}    
\begingroup%
  \makeatletter%
  \providecommand\color[2][]{%
    \errmessage{(Inkscape) Color is used for the text in Inkscape, but the package 'color.sty' is not loaded}%
    \renewcommand\color[2][]{}%
  }%
  \providecommand\transparent[1]{%
    \errmessage{(Inkscape) Transparency is used (non-zero) for the text in Inkscape, but the package 'transparent.sty' is not loaded}%
    \renewcommand\transparent[1]{}%
  }%
  \providecommand\rotatebox[2]{#2}%
  \newcommand*\fsize{\dimexpr\f@size pt\relax}%
  \newcommand*\lineheight[1]{\fontsize{\fsize}{#1\fsize}\selectfont}%
  \ifx\svgwidth\undefined%
    \setlength{\unitlength}{597.579908bp}%
    \ifx\svgscale\undefined%
      \relax%
    \else%
      \setlength{\unitlength}{\unitlength * \real{\svgscale}}%
    \fi%
  \else%
    \setlength{\unitlength}{\svgwidth}%
  \fi%
  \global\let\svgwidth\undefined%
  \global\let\svgscale\undefined%
  \makeatother%
  \begin{picture}(1,0.25034942)%
  \scriptsize
    \lineheight{1}%
    \setlength\tabcolsep{0pt}%
    \put(0,0){\includegraphics[width=\unitlength]{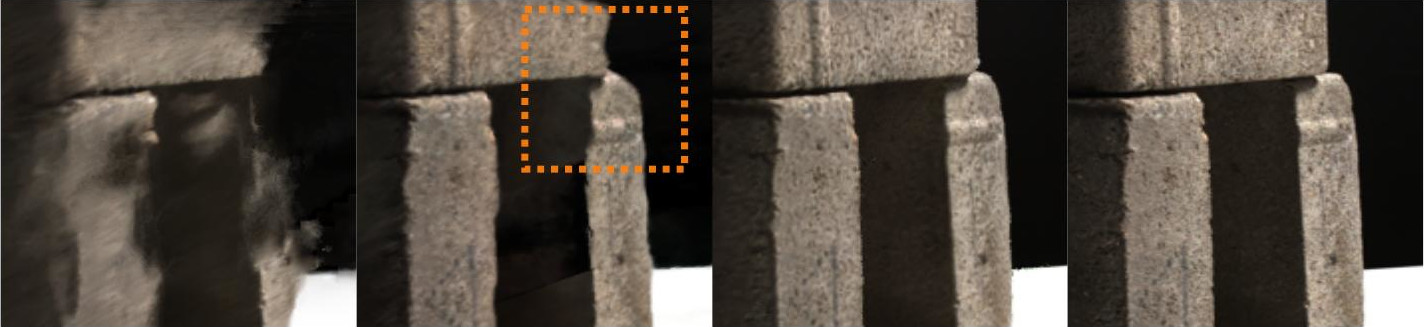}}%
    \put(0.1265783,0.2402828){\makebox(0,0)[t]{\lineheight{1.25}\smash{\begin{tabular}[t]{c}NeuRay\end{tabular}}}}%
    \put(0.37554498,0.2402828){\makebox(0,0)[t]{\lineheight{1.25}\smash{\begin{tabular}[t]{c}MVSNeRF\end{tabular}}}}%
    \put(0.62391086,0.2402828){\makebox(0,0)[t]{\lineheight{1.25}\smash{\begin{tabular}[t]{c}Ours\end{tabular}}}}%
    \put(0.87304173,0.2402828){\makebox(0,0)[t]{\lineheight{1.25}\smash{\begin{tabular}[t]{c}Ground Truth\end{tabular}}}}%
  \end{picture}%
\endgroup%

    \caption{Qualitative comparison to NeuRay~\cite{liu2022neuray} and MVSNeRF~\cite{chen:mvsnerf} on DTU~\cite{jensen2014large}.}
    \vspace{-0.35cm}
    \label{fig:qual_eval_dtu_mvsnerf_neuray}
\end{figure}
\begin{table}[h]
    \resizebox{\linewidth}{!}{%
    \begin{tabular}{l|ccccc} \toprule
        \textbf{Method}& \textbf{LPIPS} $\downarrow$& \textbf{L1} $\downarrow$& \textbf{L2} $\downarrow$& \textbf{PSNR} $\uparrow$& \textbf{SSIM} $\uparrow$\\ \midrule
        NeuRay & $0.41$ & $0.069$ & $0.017$& $19.50$& $0.65$\\
        MVSNeRF & $0.35$  & $0.059$  & $0.013$& $20.45$& $0.67$\\
        IBRNet \cite{wang2021ibrnet}& $0.40$& $0.066$& $0.017$& $19.94$& $0.65$\\
        pixelNeRF \cite{yu2020pixelnerf} & $0.38$& $0.055$& $0.011$& $20.96$& $0.67$\\
        KeypointNeRF \cite{Mihajlovic:ECCV2022}& $-$& $-$& $-$& $-$& $-$\\\midrule
        Ours w/o $\mathcal{L}_\text{vgg}$ & $0.27$& $\mathbf{0.037}$& $\mathbf{0.006}$& $\mathbf{24.14}$& $\mathbf{0.82}$\\
        Ours & $\mathbf{0.23}$& $0.039$& $0.007$& $23.44$& $0.81$\\\bottomrule
    \end{tabular}
    }
    \caption{Quantitative comparison on DTU~\cite{jensen2014large}.}
    \label{tab:quant_comparison_dtu}
\end{table}

We presented a qualitative comparison for novel view synthesis of general objects in the DTU dataset~\cite{jensen2014large} in the main paper and in \Cref{fig:dtu_qual_2}. The quantitative evaluation is provided in \Cref{tab:quant_comparison_dtu}.
Please note that KeypointNeRF~\cite{Mihajlovic:ECCV2022} cannot be applied to general objects since keypoints cannot be generalized to arbitrary objects and that we added two further baseline methods NeuRay~\cite{liu2022neuray} and MVSNeRF~\cite{chen:mvsnerf}. The qualitative comparison between DINER and NeuRay and MVSNeRF can be found in \cref{fig:qual_eval_dtu_mvsnerf_neuray}.
Our method outperforms all baseline methods by a significant margin. The improvements are even more noticeable than for the FaceScape dataset~\cite{yang2020facescape} for which we presented the quantitative results in the main paper. 
We found that while previous methods are able to learn a coarse geometry prior when applied to heads only, i.e. when trained and evaluated on FaceScape, they fail to do so for general scenes. 
As a consequence, exploiting depth information to guide the synthesis of general scenes is even more beneficial.
On the other hand, we found that adding a perceptual loss does not increase the synthesis quality as much as for FaceScape. 

\begin{figure*}[t]
\begin{center}
    \vspace{-0.5cm}
    \input{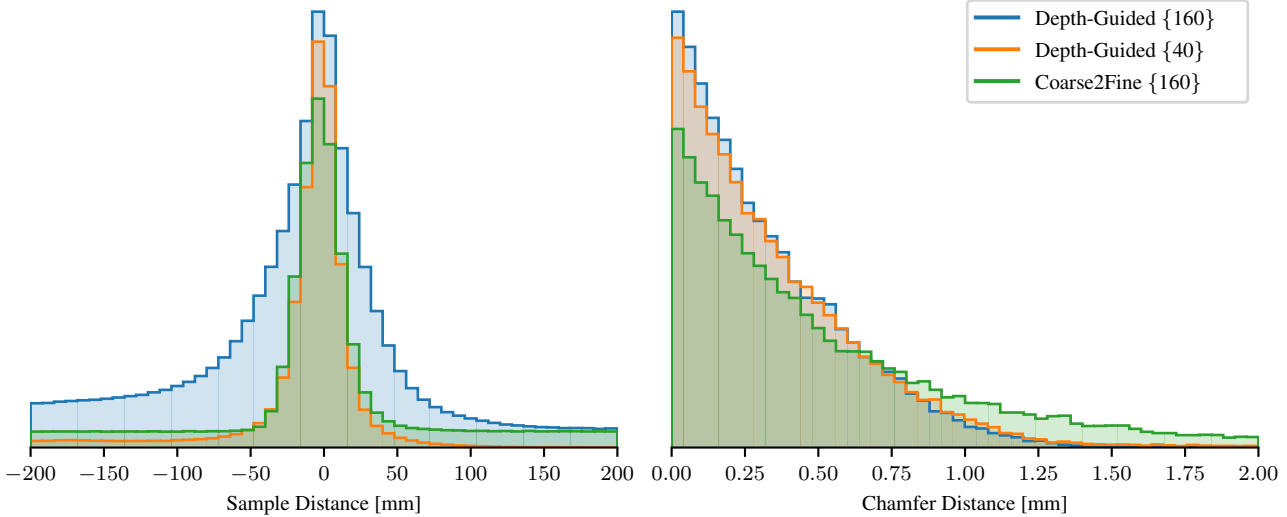}
\end{center}
\vspace{-.7cm}
\caption{Distances between sampled points and ground truth surface for depth-guided sampling and standard coarse-to-fine-based sampling as in the original NeRF paper~\cite{mildenhall2020nerf}. Curly braces indicate the number of samples per ray. Left: distances between sampling points and ground truth surface. Right: distances between ground truth surface and closest sampling point (\emph{Chamfer distance}). Depth-guided sampling effectively focuses the sampling on the ground truth surface and places samples closer to it.}
\label{fig:depth_guided_sampling_stats}
\end{figure*}

\section{Influence of Depth Accuracy}
\label{sec:depth_acc}
Since our method relies on predicted depth maps which are subject to inaccuracies,  we investigate how depth accuracy reflects on the synthesis quality. To this end, we perform a set of experiments where we train our model on the ground truth depth perturbed by Gaussian noise with varying standard deviations. \Cref{fig:depth_noise_ablation} displays the quantitative and qualitative findings. We observe that higher depth accuracy also improves the synthesis quality up until a standard deviation of 1mm. More accurate depth information does not improve synthesis quality further. We conclude that a better depth estimation network could yield an additional boost to our model's performance. 


\section{Depth-Guided Sampling}
\label{sec:eval_depth_guided_sampling}
\begin{table}[t]
\resizebox{\linewidth}{!}{%
\begin{tabular}{l|cc} \toprule
\textbf{Sampling Strategy} & Median Chamfer Dist. & Maximum Chamfer Dist. \\ \midrule
Coarse-to-Fine \{160\}   & 0.39 mm                     & 6.7 mm                   \\
Coarse-to-Fine \{40\}    & 6.4 mm                     & 55.6 mm                  \\ 
Depth-Guided \{160\} & 0.26 mm                     & 1.6 mm                   \\
Depth-Guided \{40\}  & 0.28 mm                     & 6.7 mm                   \\
\bottomrule
\end{tabular}
}
\caption{Distances between the ground truth surface and the closest sampling points (\emph{Chamfer distance}) for different sampling strategies. Curly brackets indicate the number of samples per ray. Depth-guided sampling places samples closer to the ground truth surface and focuses on these areas even if only few samples are drawn.}
\label{tab:depth_guided_sampling}
\end{table}
In this section, we analyze how depth guidance improves sampling efficiency. More specifically, we measure how close the sampled points lie around the ground truth surface. 
For this, we consider two quantities: the distances between sampled points and the ground truth surface, and the distances between the ground truth surface and closest sampling points, i.e., the Chamfer distance. \Cref{fig:depth_guided_sampling_stats} visualizes both distributions in comparison to the standard coarse-to-fine sampling strategy as introduced in the original NeRF paper~\cite{mildenhall2020nerf}. 
%
%
In \Cref{fig:depth_guided_sampling_stats}(left), we observe that coarse-to-fine sampling places a comparably small number of samples close the to ground truth surface. This is because first, a partition of the samples must be used to query the coarse MLP to find regions of interest; second, even a part of the remaining samples is used to uniformly query the space which leads to long, non-vanishing tails in the distance distributions. 
As a consequence of the low sample density around the ground truth surface, we observe fewer surface points with small Chamfer distances in \Cref{fig:depth_guided_sampling_stats}(right) and a comparatively high median Chamfer distance in \Cref{tab:depth_guided_sampling}. 
In contrast, depth-guided sampling with the same number of points per ray places more samples closer to the ground truth surface (see \Cref{fig:depth_guided_sampling_stats}), which reduces the median Chamfer distance by $33\%$ and the maximum Chamfer distance by a factor of $4$ (see \Cref{tab:depth_guided_sampling}). 
\begin{figure}[t]
    \centering    
        \def\svgwidth{\linewidth}
\begingroup%
  \makeatletter%
  \providecommand\color[2][]{%
    \errmessage{(Inkscape) Color is used for the text in Inkscape, but the package 'color.sty' is not loaded}%
    \renewcommand\color[2][]{}%
  }%
  \providecommand\transparent[1]{%
    \errmessage{(Inkscape) Transparency is used (non-zero) for the text in Inkscape, but the package 'transparent.sty' is not loaded}%
    \renewcommand\transparent[1]{}%
  }%
  \providecommand\rotatebox[2]{#2}%
  \newcommand*\fsize{\dimexpr\f@size pt\relax}%
  \newcommand*\lineheight[1]{\fontsize{\fsize}{#1\fsize}\selectfont}%
  \ifx\svgwidth\undefined%
    \setlength{\unitlength}{562.49987024bp}%
    \ifx\svgscale\undefined%
      \relax%
    \else%
      \setlength{\unitlength}{\unitlength * \real{\svgscale}}%
    \fi%
  \else%
    \setlength{\unitlength}{\svgwidth}%
  \fi%
  \global\let\svgwidth\undefined%
  \global\let\svgscale\undefined%
  \makeatother%
  \begin{picture}(1,0.43878909)%
    \lineheight{1}%
    \setlength\tabcolsep{0pt}%
    \put(0,0){\includegraphics[width=\unitlength,page=1]{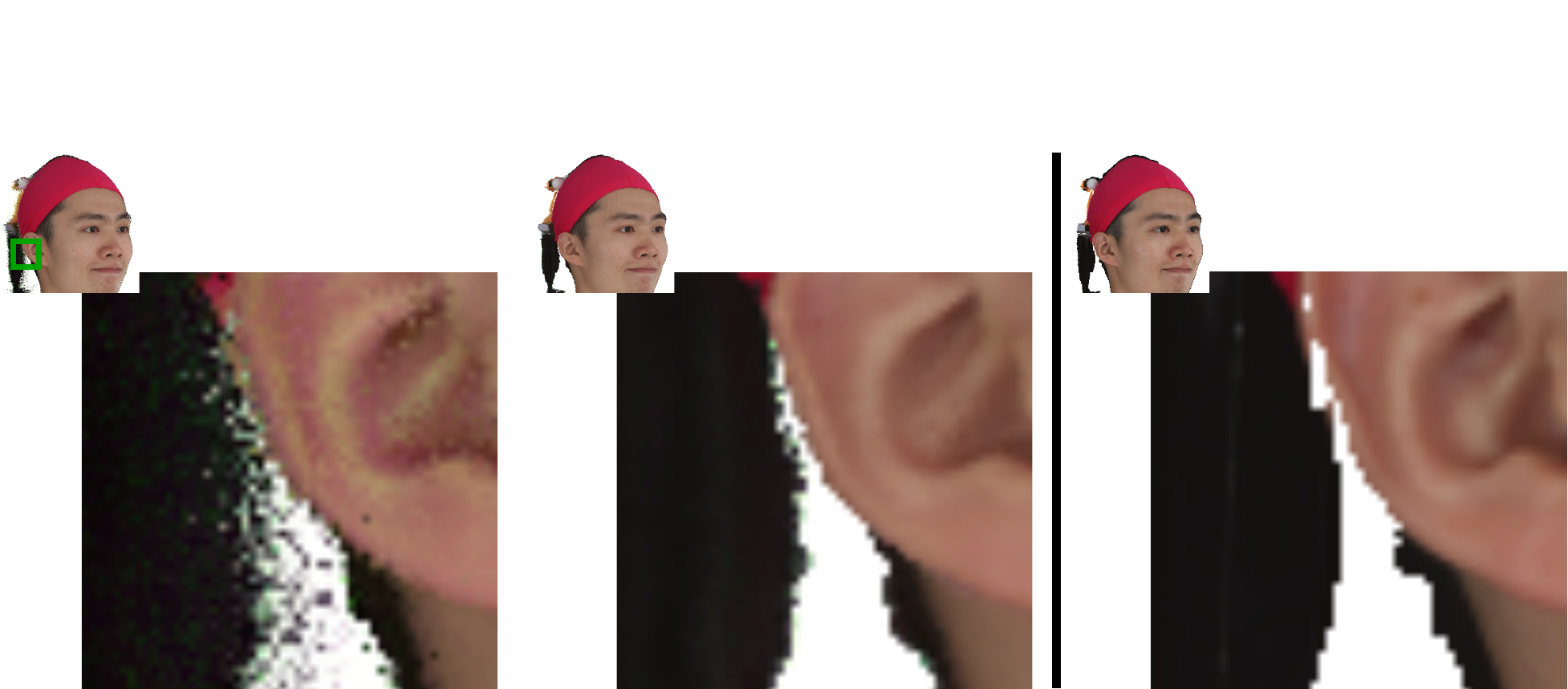}}%
    \put(0.16726197,0.42798354){\color[rgb]{0,0,0}\makebox(0,0)[t]{\lineheight{1.25}\smash{\begin{tabular}[t]{c}Ours\\(coarse2fine)\end{tabular}}}}%
    \put(0.49096007,0.42798354){\color[rgb]{0,0,0}\makebox(0,0)[t]{\lineheight{1.25}\smash{\begin{tabular}[t]{c}Ours\\(depth guided)\end{tabular}}}}%
    \put(0.86714833,0.42798354){\color[rgb]{0,0,0}\makebox(0,0)[t]{\lineheight{1.25}\smash{\begin{tabular}[t]{c}Ground\\Truth\end{tabular}}}}%
    \put(0,0){\includegraphics[width=\unitlength,page=2]{ours_coarse2fine.pdf}}%
  \end{picture}%
\endgroup%

    \caption{Qualitative comparison of sampling strategies. Both models sample only 40 points per ray and were trained with batch size 4. Depth guidance improves sampling efficiency and solves artifacts around thin surfaces.}
    \label{fig:coarse2fine_40points}
\end{figure}
Note that depth-guided sampling does not require querying a coarse MLP and, therefore, more samples contribute directly to the final output color. 
Even when we reduce the number of samples by a factor of 4, \Cref{fig:depth_guided_sampling_stats}(left) shows that depth-guided sampling focuses on areas close to the ground truth surfaces and predominantly minimizes the tails of the distance distribution, i.e., drops samples that lie far away from the surface. 
As a consequence, compared to standard coarse-to-fine sampling with 4 times more samples, we observe a significantly improved median Chamfer distance (see \Cref{tab:depth_guided_sampling}). 
In contrast, when cutting the number of samples per ray for standard coarse-to-fine sampling, we observe significantly degraded Chamfer distances. \Cref{fig:coarse2fine_40points} demonstrates that this results in severe artifacts around thin surfaces during novel view synthesis. We conclude that only depth-guided sampling allows us to cut the number of sampled points per ray by a factor of 4 without introducing artifacts. This in turn allows us to increase the batch size during training from 1 to 4 without changing hardware requirements which we found to improve model performance.


\section{Ethical Considerations}
\label{sec:ethical_considerations}
Our method reconstructs a volumetric representation of a subject or general objects from sparse color camera inputs.
Since this volumetric representation does only allow for novel view synthesis, there is no immediate risk of misuse, such as deep fakes.
As no personalized avatar is reconstructed, a potential immersive telepresence application does not need to store person-specific information.
We train the method on FaceScape~\cite{yang2020facescape} which is not a balanced face dataset and is biased towards the local population.
However, in the main paper, we show that DINER generalizes well to subjects of unseen ethnicities and therefore rules out discrimination against underrepresented minorities. 

The human data used in this study is based on the FaceScape dataset with the consent of the subjects to be used for research.
Four subjects agreed to be displayed in publications and presentations; these subjects are the test set.

\clearpage
{\small
\bibliographystyle{ieee_fullname}
\bibliography{egbib}

\begin{thebibliography}{10}\itemsep=-1pt

\bibitem{Barron2021}
Jonathan~T. Barron, Ben Mildenhall, Matthew Tancik, Peter Hedman, Ricardo
  Martin-Brualla, and Pratul~P. Srinivasan.
\newblock Mip-nerf: A multiscale representation for anti-aliasing neural
  radiance fields.
\newblock Mar. 2021.

\bibitem{buehler2021varitex}
Marcel~C. Buehler, Abhimitra Meka, Gengyan Li, Thabo Beeler, and Otmar
  Hilliges.
\newblock Varitex: Variational neural face textures.
\newblock In {\em Proceedings of the IEEE/CVF International Conference on
  Computer Vision}, 2021.

\bibitem{authenticvolumetricavatars}
Chen Cao, Tomas Simon, Jin~Kyu Kim, Gabe Schwartz, Michael Zollhoefer, Shunsuke
  Saito, Stephen Lombardi, Shih en Wei, Danielle Belko, Shoou i Yu, Yaser
  Sheikh, and Jason Saragih.
\newblock Authentic volumetric avatars from a phone scan.
\newblock {\em SIGGRAPH}, 2022.

\bibitem{Chan2021}
Eric~R. Chan, Connor~Z. Lin, Matthew~A. Chan, Koki Nagano, Boxiao Pan,
  Shalini~De Mello, Orazio Gallo, Leonidas Guibas, Jonathan Tremblay, Sameh
  Khamis, Tero Karras, and Gordon Wetzstein.
\newblock Eg3d: Efficient geometry-aware {3D} generative adversarial networks.
\newblock In {\em arXiv}, 2021.

\bibitem{chen2022tensorf}
Anpei Chen, Zexiang Xu, Andreas Geiger, Jingyi Yu, and Hao Su.
\newblock Tensorf: Tensorial radiance fields.
\newblock {\em arXiv preprint arXiv:2203.09517}, 2022.

\bibitem{chen:mvsnerf}
Anpei Chen, Zexiang Xu, Fuqiang Zhao, Xiaoshuai Zhang, Fanbo Xiang, Jingyi Yu,
  and Hao Su.
\newblock Mvsnerf: Fast generalizable radiance field reconstruction from
  multi-view stereo, 2021.

\bibitem{kangle2021dsnerf}
Kangle Deng, Andrew Liu, Jun-Yan Zhu, and Deva Ramanan.
\newblock Depth-supervised {NeRF}: Fewer views and faster training for free.
\newblock In {\em Proceedings of the IEEE/CVF Conference on Computer Vision and
  Pattern Recognition (CVPR)}, June 2022.

\bibitem{ding2022transmvsnet}
Yikang Ding, Wentao Yuan, Qingtian Zhu, Haotian Zhang, Xiangyue Liu, Yuanjiang
  Wang, and Xiao Liu.
\newblock Transmvsnet: Global context-aware multi-view stereo network with
  transformers.
\newblock In {\em Proceedings of the IEEE/CVF Conference on Computer Vision and
  Pattern Recognition}, pages 8585--8594, 2022.

\bibitem{megaportraits}
Nikita Drobyshev, Jenya Chelishev, Taras Khakhulin, Aleksei Ivakhnenko, Victor
  Lempitsky, and Egor Zakharov.
\newblock Megaportraits: One-shot megapixel neural head avatars, 2022.

\bibitem{DECA:Siggraph2021}
Yao Feng, Haiwen Feng, Michael~J. Black, and Timo Bolkart.
\newblock Learning an animatable detailed {3D} face model from in-the-wild
  images.
\newblock volume~40, 2021.

\bibitem{Gafni2020}
Guy Gafni, Justus Thies, Michael Zollhoefer, and Matthias Nießner.
\newblock Dynamic neural radiance fields for monocular 4d facial avatar
  reconstruction.
\newblock Dec. 2020.

\bibitem{voltemorph}
Stephan~J. Garbin, Marek Kowalski, Virginia Estellers, Stanislaw Szymanowicz,
  Shideh Rezaeifar, Jingjing Shen, Matthew Johnson, and Julien Valentin.
\newblock Voltemorph: Realtime, controllable and generalisable animation of
  volumetric representations, 2022.

\bibitem{fastnerf}
Stephan~J. Garbin, Marek Kowalski, Matthew Johnson, Jamie Shotton, and Julien
  Valentin.
\newblock Fastnerf: High-fidelity neural rendering at 200fps, 2021.

\bibitem{DBLP:journals/corr/abs-2112-01554}
Philip{-}William Grassal, Malte Prinzler, Titus Leistner, Carsten Rother,
  Matthias Nie{\ss}ner, and Justus Thies.
\newblock Neural head avatars from monocular {RGB} videos.
\newblock {\em CoRR}, abs/2112.01554, 2021.

\bibitem{Grigorev_2021_CVPR_stylepeople}
Artur Grigorev, Karim Iskakov, Anastasia Ianina, Renat Bashirov, Ilya
  Zakharkin, Alexander Vakhitov, and Victor Lempitsky.
\newblock Stylepeople: A generative model of fullbody human avatars.
\newblock In {\em Proceedings of the IEEE/CVF Conference on Computer Vision and
  Pattern Recognition (CVPR)}, pages 5151--5160, June 2021.

\bibitem{headnerf}
Yang Hong, Bo Peng, Haiyao Xiao, Ligang Liu, and Juyong Zhang.
\newblock Headnerf: A real-time nerf-based parametric head model, 2021.

\bibitem{Hu2022}
Tao Hu, Shu Liui, Lun Chen, Tiancheng Shen, and Jiaya Jia.
\newblock Efficientnerf – efficient neural radiance fields.
\newblock {\em CVPR}, 2022.

\bibitem{jensen2014large}
Rasmus Jensen, Anders Dahl, George Vogiatzis, Engil Tola, and Henrik Aan{\ae}s.
\newblock Dtu: Large scale multi-view stereopsis evaluation.
\newblock In {\em 2014 IEEE Conference on Computer Vision and Pattern
  Recognition}, pages 406--413. IEEE, 2014.

\bibitem{kania2021conerf}
Kacper Kania, Kwang~Moo Yi, Marek Kowalski, Tomasz Trzci{\'n}ski, and Andrea
  Tagliasacchi.
\newblock Conerf: Controllable neural radiance fields, 2021.

\bibitem{oneshotmeshavatars}
Taras Khakhulin, Vanessa Sklyarova, Victor Lempitsky, and Egor Zakharov.
\newblock Realistic one-shot mesh-based head avatars, 2022.

\bibitem{kim2018deep}
Hyeongwoo Kim, Pablo Garrido, Ayush Tewari, Weipeng Xu, Justus Thies, Matthias
  Nie{\ss}ner, Patrick P{\'e}rez, Christian Richardt, Michael Zoll{\"o}fer, and
  Christian Theobalt.
\newblock Deep video portraits.
\newblock {\em ACM Transactions on Graphics (TOG)}, 37(4):163, 2018.

\bibitem{Efficientnet}
Haotong Lin, Sida Peng, Zhen Xu, Hujun Bao, and Xiaowei Zhou.
\newblock Efficient neural radiance fields with learned depth-guided sampling,
  2021.

\bibitem{liu2022neuray}
Yuan Liu, Sida Peng, Lingjie Liu, Qianqian Wang, Peng Wang, Christian Theobalt,
  Xiaowei Zhou, and Wenping Wang.
\newblock Neural rays for occlusion-aware image-based rendering.
\newblock In {\em CVPR}, 2022.

\bibitem{Lombardi:2019}
Stephen Lombardi, Tomas Simon, Jason Saragih, Gabriel Schwartz, Andreas
  Lehrmann, and Yaser Sheikh.
\newblock Neural volumes: Learning dynamic renderable volumes from images.
\newblock {\em ACM Trans. Graph.}, 38(4):65:1--65:14, July 2019.

\bibitem{lombardi2021mixture}
Stephen Lombardi, Tomas Simon, Gabriel Schwartz, Michael Zollhoefer, Yaser
  Sheikh, and Jason Saragih.
\newblock Mixture of volumetric primitives for efficient neural rendering,
  2021.

\bibitem{Mihajlovic:ECCV2022}
Marko Mihajlovic, Aayush Bansal, Michael Zollhoefer, Siyu Tang, and Shunsuke
  Saito.
\newblock {KeypointNeRF}: Generalizing image-based volumetric avatars using
  relative spatial encoding of keypoints.
\newblock In {\em European conference on computer vision}, 2022.

\bibitem{Mildenhall2022CVPR}
Ben Mildenhall, Peter Hedman, Ricardo Martin-Brualla, Pratul~P. Srinivasan, and
  Jonathan~T. Barron.
\newblock Nerf in the dark: High dynamic range view synthesis from noisy raw
  images.
\newblock In {\em Proceedings of the IEEE/CVF Conference on Computer Vision and
  Pattern Recognition (CVPR)}, pages 16190--16199, June 2022.

\bibitem{mildenhall2020nerf}
Ben Mildenhall, Pratul~P Srinivasan, Matthew Tancik, Jonathan~T Barron, Ravi
  Ramamoorthi, and Ren Ng.
\newblock Nerf: Representing scenes as neural radiance fields for view
  synthesis.
\newblock In {\em European conference on computer vision}, pages 405--421.
  Springer, 2020.

\bibitem{mueller2022instant}
Thomas Mueller, Alex Evans, Christoph Schied, and Alexander Keller.
\newblock Instant neural graphics primitives with a multiresolution hash
  encoding, 2022.

\bibitem{Niemeyer2021Regnerf}
Michael Niemeyer, Jonathan~T. Barron, Ben Mildenhall, Mehdi S.~M. Sajjadi,
  Andreas Geiger, and Noha Radwan.
\newblock Regnerf: Regularizing neural radiance fields for view synthesis from
  sparse inputs.
\newblock In {\em Proc. IEEE Conf. on Computer Vision and Pattern Recognition
  (CVPR)}, 2022.

\bibitem{park2021nerfies}
Keunhong Park, Utkarsh Sinha, Jonathan~T Barron, Sofien Bouaziz, Dan~B Goldman,
  Steven~M Seitz, and Ricardo Martin-Brualla.
\newblock Nerfies: Deformable neural radiance fields.
\newblock In {\em Proceedings of the IEEE/CVF International Conference on
  Computer Vision}, pages 5865--5874, 2021.

\bibitem{Park2021}
Keunhong Park, Utkarsh Sinha, Peter Hedman, Jonathan~T. Barron, Sofien Bouaziz,
  Dan~B Goldman, Ricardo Martin-Brualla, and Steven~M. Seitz.
\newblock Hypernerf: A higher-dimensional representation for topologically
  varying neural radiance fields.
\newblock {\em ACM Trans. Graph.}, 40(6), dec 2021.

\bibitem{peng2021neuralbody}
Sida Peng, Yuanqing Zhang, Yinghao Xu, Qianqian Wang, Qing Shuai, Hujun Bao,
  and Xiaowei Zhou.
\newblock Neural body: Implicit neural representations with structured latent
  codes for novel view synthesis of dynamic humans.
\newblock In {\em CVPR}, 2021.

\bibitem{Pumarola2020}
Albert Pumarola, Enric Corona, Gerard Pons-Moll, and Francesc Moreno-Noguer.
\newblock D-nerf: Neural radiance fields for dynamic scenes.
\newblock jun 2021.

\bibitem{Ramon_2021_ICCV}
Eduard Ramon, Gil Triginer, Janna Escur, Albert Pumarola, Jaime Garcia, Xavier
  Gir\'o-i Nieto, and Francesc Moreno-Noguer.
\newblock H3d-net: Few-shot high-fidelity 3d head reconstruction.
\newblock In {\em Proceedings of the IEEE/CVF International Conference on
  Computer Vision (ICCV)}, pages 5620--5629, October 2021.

\bibitem{rematas2021urban}
Konstantinos Rematas, Andrew Liu, Pratul~P Srinivasan, Jonathan~T Barron,
  Andrea Tagliasacchi, Thomas Funkhouser, and Vittorio Ferrari.
\newblock Urban radiance fields.
\newblock {\em arXiv preprint arXiv:2111.14643}, 2021.

\bibitem{roessle2022depthpriorsnerf}
Barbara Roessle, Jonathan~T. Barron, Ben Mildenhall, Pratul~P. Srinivasan, and
  Matthias Nie{\ss}ner.
\newblock Dense depth priors for neural radiance fields from sparse input
  views.
\newblock In {\em Proceedings of the IEEE/CVF Conference on Computer Vision and
  Pattern Recognition (CVPR)}, June 2022.

\bibitem{rosu2022neuralstrands}
Radu~Alexandru Rosu, Shunsuke Saito, Ziyan Wang, Chenglei Wu, Sven Behnke, and
  Giljoo Nam.
\newblock Neural strands: Learning hair geometry and appearance from multi-view
  images.
\newblock {\em ECCV}, 2022.

\bibitem{srt22}
Mehdi S.~M. Sajjadi, Henning Meyer, Etienne Pot, Urs Bergmann, Klaus Greff,
  Noha Radwan, Suhani Vora, Mario Lucic, Daniel Duckworth, Alexey Dosovitskiy,
  Jakob Uszkoreit, Thomas Funkhouser, and Andrea Tagliasacchi.
\newblock {Scene Representation Transformer: Geometry-Free Novel View Synthesis
  Through Set-Latent Scene Representations}.
\newblock {\em {CVPR}}, 2022.

\bibitem{Siarohin2019}
Aliaksandr Siarohin.
\newblock First order motion model for image animation.
\newblock {\em Advances in Neural Information Processing Systems},
  32:7137--7147, jun 2019.

\bibitem{vggloss}
Karen Simonyan and Andrew Zisserman.
\newblock Very deep convolutional networks for large-scale image recognition,
  2014.

\bibitem{suhail2022generalizable}
Mohammed Suhail, Carlos Esteves, Leonid Sigal, and Ameesh Makadia.
\newblock Generalizable patch-based neural rendering.
\newblock In {\em European Conference on Computer Vision}. Springer, 2022.

\bibitem{tancik2022blocknerf}
Matthew Tancik, Vincent Casser, Xinchen Yan, Sabeek Pradhan, Ben Mildenhall,
  Pratul Srinivasan, Jonathan~T. Barron, and Henrik Kretzschmar.
\newblock {Block-NeRF}: Scalable large scene neural view synthesis.
\newblock {\em arXiv}, 2022.

\bibitem{tewari2022advances}
Ayush Tewari, Justus Thies, Ben Mildenhall, Pratul Srinivasan, Edgar Tretschk,
  Yifan Wang, Christoph Lassner, Vincent Sitzmann, Ricardo Martin-Brualla,
  Stephen Lombardi, Tomas Simon, Christian Theobalt, Matthias Niessner,
  Jonathan~T. Barron, Gordon Wetzstein, Michael Zollhoefer, and Vladislav
  Golyanik.
\newblock Advances in neural rendering.
\newblock 2022.

\bibitem{tewari2020neuralrendering}
J. Thies, A. Tewari, O. Fried, V. Sitzmann, S. Lombardi, K. Sunkavalli, R.
  Martin-Brualla, T. Simon, J. Saragih, M. Nie{\ss}ner, R. Pandey, S. Fanello,
  G. Wetzstein, J.-Y. Zhu, C. Theobalt, M. Agrawala, E. Shechtman, D.~B
  Goldman, and M. Zollh{\"o}fer.
\newblock State of the art on neural rendering.
\newblock {\em EG}, 2020.

\bibitem{thies2019deferred}
Justus Thies, Michael Zollh{\"o}fer, and Matthias Nie{\ss}ner.
\newblock Deferred neural rendering: Image synthesis using neural textures.
\newblock {\em ACM Transactions on Graphics (TOG)}, 38(4):1--12, 2019.

\bibitem{wang2021ibrnet}
Qianqian Wang, Zhicheng Wang, Kyle Genova, Pratul Srinivasan, Howard Zhou,
  Jonathan~T. Barron, Ricardo Martin-Brualla, Noah Snavely, and Thomas
  Funkhouser.
\newblock Ibrnet: Learning multi-view image-based rendering.
\newblock In {\em CVPR}, 2021.

\bibitem{ARAH:ECCV:2022}
Shaofei Wang, Katja Schwarz, Andreas Geiger, and Siyu Tang.
\newblock Arah: Animatable volume rendering of articulated human sdfs.
\newblock In {\em European Conference on Computer Vision}, 2022.

\bibitem{wang2021facevid2vid}
Ting-Chun Wang, Arun Mallya, and Ming-Yu Liu.
\newblock One-shot free-view neural talking-head synthesis for video
  conferencing.
\newblock In {\em Proceedings of the IEEE Conference on Computer Vision and
  Pattern Recognition}, 2021.

\bibitem{Wang_2021_CVPR}
Ziyan Wang, Timur Bagautdinov, Stephen Lombardi, Tomas Simon, Jason Saragih,
  Jessica Hodgins, and Michael Zollhofer.
\newblock Learning compositional radiance fields of dynamic human heads.
\newblock In {\em Proceedings of the IEEE/CVF Conference on Computer Vision and
  Pattern Recognition (CVPR)}, pages 5704--5713, June 2021.

\bibitem{wang2021hvh}
Ziyan Wang, Giljoo Nam, Tuur Stuyck, Stephen Lombardi, Michael Zollhoefer,
  Jessica Hodgins, and Christoph Lassner.
\newblock Hvh: Learning a hybrid neural volumetric representation for dynamic
  hair performance capture, 2021.

\bibitem{weng2022humannerf}
Chung-Yi Weng, Brian Curless, Pratul~P. Srinivasan, Jonathan~T. Barron, and Ira
  Kemelmacher-Shlizerman.
\newblock Humannerf: Free-viewpoint rendering of moving people from monocular
  video.
\newblock {\em arXiv preprint arXiv:2201.04127}, 2022.

\bibitem{wuu2022multiface}
Cheng-hsin Wuu, Ningyuan Zheng, Scott Ardisson, Rohan Bali, Danielle Belko,
  Eric Brockmeyer, Lucas Evans, Timothy Godisart, Hyowon Ha, Alexander Hypes,
  Taylor Koska, Steven Krenn, Stephen Lombardi, Xiaomin Luo, Kevyn McPhail,
  Laura Millerschoen, Michal Perdoch, Mark Pitts, Alexander Richard, Jason
  Saragih, Junko Saragih, Takaaki Shiratori, Tomas Simon, Matt Stewart, Autumn
  Trimble, Xinshuo Weng, David Whitewolf, Chenglei Wu, Shoou-I Yu, and Yaser
  Sheikh.
\newblock Multiface: A dataset for neural face rendering.
\newblock In {\em arXiv}, 2022.

\bibitem{Xu2022}
Qiangeng Xu, Zexiang Xu, Julien Philip, Sai Bi, Zhixin Shu, Kalyan Sunkavalli,
  and Ulrich Neumann.
\newblock Point-nerf: Point-based neural radiance fields.
\newblock Jan. 2022.

\bibitem{yang2020facescape}
Haotian Yang, Hao Zhu, Yanru Wang, Mingkai Huang, Qiu Shen, Ruigang Yang, and
  Xun Cao.
\newblock Facescape: A large-scale high quality 3d face dataset and detailed
  riggable 3d face prediction.
\newblock In {\em IEEE/CVF Conference on Computer Vision and Pattern
  Recognition (CVPR)}, June 2020.

\bibitem{yu2021plenoctrees}
Alex Yu, Ruilong Li, Matthew Tancik, Hao Li, Ren Ng, and Angjoo Kanazawa.
\newblock {PlenOctrees} for real-time rendering of neural radiance fields.
\newblock In {\em ICCV}, 2021.

\bibitem{yu2020pixelnerf}
Alex Yu, Vickie Ye, Matthew Tancik, and Angjoo Kanazawa.
\newblock {pixelNeRF}: Neural radiance fields from one or few images.
\newblock In {\em CVPR}, 2021.

\bibitem{Zakharov20}
Egor Zakharov, Aleksei Ivakhnenko, Aliaksandra Shysheya, and Victor Lempitsky.
\newblock Fast bi-layer neural synthesis of one-shot realistic head avatars.
\newblock In {\em European Conference of Computer vision (ECCV)}, pages
  524--540, August 2020.

\bibitem{Zheng2021}
Yufeng Zheng, Victoria~Fernández Abrevaya, Xu Chen, Marcel~C. Buehler,
  Michael~J. Black, and Otmar Hilliges.
\newblock I m avatar: Implicit morphable head avatars from videos.
\newblock Dec. 2021.

\bibitem{zollhoefer2018facestar}
Michael Zollh{\"o}fer, Justus Thies, Darek Bradley, Pablo Garrido, Thabo
  Beeler, Patrick P{\'e}erez, Marc Stamminger, Matthias Nie{\ss}ner, and
  Christian Theobalt.
\newblock State of the art on monocular 3d face reconstruction, tracking, and
  applications.
\newblock 2018.

\end{thebibliography}
}
\end{document}